%% file: main.tex
\pdfoutput=1

\documentclass[11pt]{article}

\usepackage[final]{acl}

\usepackage{times}
\usepackage{latexsym}

\usepackage{blindtext} 
\usepackage{array}\newcolumntype{P}[1]{>{\centering\arraybackslash}p{#1}}
\usepackage{makecell}
\usepackage{graphicx}
\usepackage{amsmath}
\usepackage{amssymb}
\usepackage{pifont}
\usepackage{booktabs}
\usepackage{enumitem}
\usepackage{soul}
\sethlcolor{Lavender}
\usepackage{moresize}
\usepackage{multicol}
\usepackage{multirow}
\usepackage[export]{adjustbox}

\usepackage[T1]{fontenc}

\usepackage[most]{tcolorbox}
\newtcolorbox{mybox}[1]{%
  colback=orange!5!white,colframe=orange!75!black,
  fonttitle=\bfseries,
  title=#1}

\newtcolorbox{examplebox}[3]{%
  colback=#3!15!white,colframe=#3!85!black,
  title=\textbf{Example #1:} #2,
  left=0pt,right=0pt,top=0pt,bottom=0pt,fonttitle=\color{black}}

\usepackage{float}

\newcommand{\cmark}{\ding{51}}%
\newcommand{\xmark}{\ding{55}}%

\definecolor{pred}{HTML}{FFBBBB}
\definecolor{pgreen}{HTML}{BBFFBB}
\definecolor{pblue}{HTML}{BBBBFF}
\definecolor{pcyan}{HTML}{BBFFFF}
\definecolor{pyellow}{HTML}{FFFFBB}
\definecolor{pblack}{HTML}{AAAAAA}
\definecolor{pwhite}{HTML}{FFFFFF}
\definecolor{darkgreen}{HTML}{00AA00}
\definecolor{darkred}{HTML}{AA0000}

\title{Transparent and Coherent Procedural Mistake Detection}

\author{Shane Storks \hspace{20pt} Itamar Bar-Yossef \hspace{20pt} Yayuan Li \hspace{20pt}  Zheyuan Zhang \\ \hspace{20pt} \textbf{Jason J. Corso} \hspace{20pt} \textbf{Joyce Chai} \\
         University of Michigan, Ann Arbor, Michigan, USA \\
         \{\texttt{sstorks}, \texttt{itamarby}, \texttt{yayuanli}, \texttt{zheyuan}, \texttt{jjcorso}, \texttt{chaijy}\}\texttt{@umich.edu}}

\begin{document}
\maketitle
\begin{abstract}
Procedural mistake detection (PMD) is a challenging problem of classifying whether a human user (observed through egocentric video) has successfully executed a task (specified by a procedural text). Despite significant recent efforts, machine performance in the wild remains nonviable, and the reasoning processes underlying this performance are opaque. As such, we extend PMD to require generating visual self-dialog rationales to inform decisions. Given the impressive, mature image understanding capabilities observed in recent vision-and-language models (VLMs), we curate a suitable benchmark dataset for PMD based on individual frames. As our reformulation enables unprecedented transparency, we leverage a natural language inference (NLI) model to formulate two automated metrics for the coherence of generated rationales. We establish baselines for this reframed task, showing that VLMs struggle off-the-shelf, but with some trade-offs, their accuracy, coherence, and efficiency can be improved by incorporating these metrics into common inference and fine-tuning methods. Lastly, our multi-faceted metrics visualize common outcomes, highlighting areas for further improvement.
\end{abstract}

\input{latex/sections/1-intro}

\input{latex/sections/2-problem}

\input{latex/sections/3-methods}

\input{latex/sections/4-results}

\input{latex/sections/5-related-work}
\input{latex/sections/6-conclusion}
\input{latex/sections/7-etc}

\bibliography{latex/references/references_1,latex/references/references_2,latex/references/references_3_4,latex/references/references_6,latex/references/references_7,latex/references/references_8,latex/references/references_my_papers}

\appendix
\input{latex/sections/8-apx}

\end{document}

%% file: latex/sections/1-intro.tex
\begin{figure*}[!ht]
    \centering
    \includegraphics[width=0.88\linewidth]{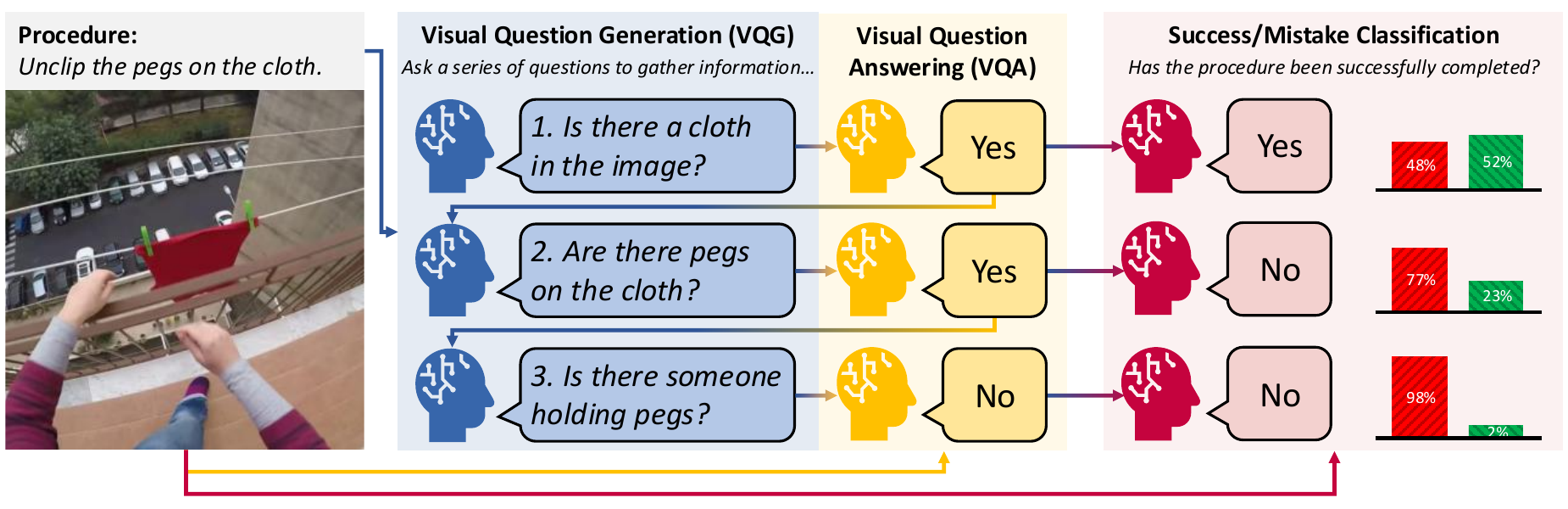}
    \vspace{-6pt}
    \caption{To reason through the complex task of procedural mistake detection (PMD), vision-and-language models (VLMs) are conditioned to gather visual evidence through an iterative self-dialog to rationalize their final decision.}
    \label{fig: self dialog}
\end{figure*}

\section{Introduction}\label{sec:intro}

The problem of interactive task guidance has recently attracted attention in AI research \cite{bao-etal-2023-foundation,HoloAssist2023,peddi2023captaincook4d,bohus2024sigma}, stemming from significant efforts to build and learn from large-scale procedural video datasets \cite{Zhou_Xu_Corso_2018,Damen2018EPICKITCHENS,miech19howto100m,Ego4D2022CVPR}. 
A successful task guidance agent can observe a human user through video and guide them to complete a task through language interaction. A key component of such an agent is \textbf{procedural mistake detection (PMD)}: the ability to detect when the user's actions deviate from a procedural text, e.g., a recipe or instruction manual. 
To achieve this, a system must apply physical and procedural commonsense knowledge to anticipate success conditions for the text, then extract relevant state information from the visual scene to verify them.

Prior work in PMD has explored a mix of specialized (primarily vision-based) classifiers \cite{sener2022assembly101,HoloAssist2023,peddi2023captaincook4d} as well as foundational language models (LMs) and vision-and-language models (VLMs) \cite{pmlr-v232-du23b,bao-etal-2023-foundation,peddi2023captaincook4d,Flaborea_2024_CVPR}, but this problem has proven difficult, and current approaches fail to achieve accuracy above chance in zero-shot PMD in the wild. Qualitatively, \citet{bao-etal-2023-foundation} found that while the web-scale multimodal pre-training of foundational VLMs enables coverage of a wide variety of procedures, they often produce noisy, vague, or otherwise insufficient information from visual scenes to facilitate PMD. 
This capability to extract and reason over key task-relevant visual information is crucial to PMD, but prior work has largely overlooked it, instead targeting binary or categorical classification tasks in their system design and quantitative evaluations. 
Consequently, the reasons for VLMs' decisions are opaque, hindering practical use\footnote{Since VLMs are unreliable mistake detectors, transparency is crucial for the user to understand the system's reasoning process and act on or disregard it accordingly.} and continued improvement.


To promote the development of PMD systems with transparent and justified decisions, we propose a reformulated problem of \textbf{coherent PMD}: given a procedural text and egocentric video frame, VLMs must not only classify whether a mistake has occurred, but also support this decision with a \textit{rationale} consisting of evidence from the visual scene. As shown in Figure~\ref{fig: self dialog}, this rationale takes the form of an iterative \textit{self-dialog} of generated questions and yes-no answers which directly condition classification.\footnote{This choice of rationale resembles visual question answering (VQA; \citealp{Antol_2015_ICCV}) and visual dialog \cite{Das_2017_CVPR}, long-studied multimodal tasks that VLMs excel at \cite{instructblip,liu2023llava,dubey2024llama3herdmodels}.} Since recent VLMs struggle to extract detailed, temporally coherent information from videos, but have exhibited more mature image understanding capabilities, we curate an approachable large-scale dataset for PMD based on individual video frames annotated in Ego4D~\cite{Ego4D2022CVPR}. We define two metrics for the coherence of generated rationales based on a natural language inference (NLI) model. To lay a foundation for research in coherent PMD, we establish baselines by exploring three natural interventions to VLMs: (1) we use our metrics to re-rank candidate questions generated by VLMs, (2) we harness VLMs' in-context learning capability to generate additional candidate questions based on human-written examples, and (3) we use our metrics to fine-tune VLMs to generate more coherent questions. Our results show that while VLMs struggle off-the-shelf, these interventions can improve VLMs' accuracy, coherence, and rationale generation efficiency, albeit creating tradeoffs between these aspects. We lastly show how our multi-faceted metrics visualize common outcomes in coherent PMD (e.g., unjustified decisions, object hallucination, and more), enabling fine-grained evaluation and identification of areas for future improvement.

%% file: latex/sections/2-problem.tex
\begin{figure*}
    \centering
    \includegraphics[width=0.95\linewidth]{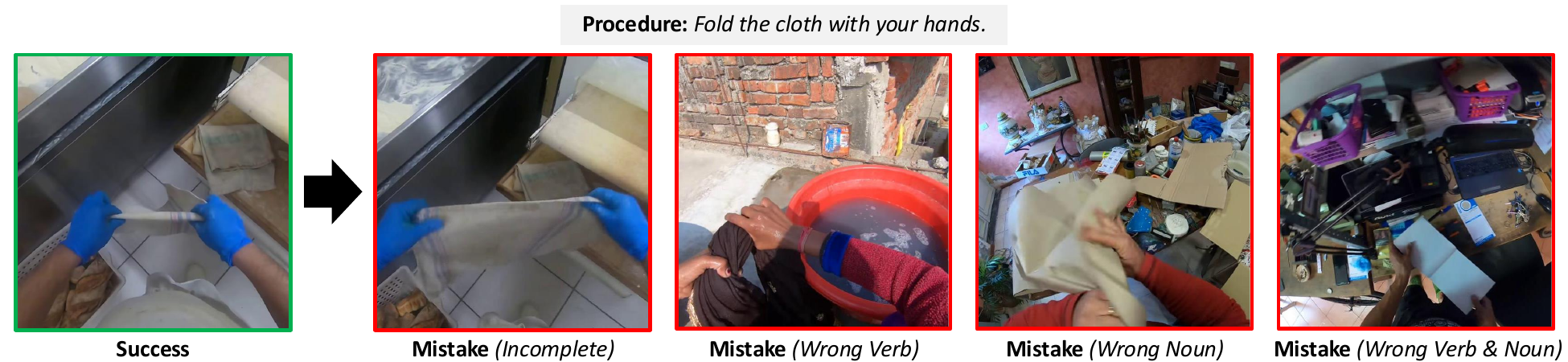}
    
    \vspace{-6pt}
    
    \caption{Selected examples from Ego4D~\cite{Ego4D2022CVPR} for Procedural Mistake Detection (Ego4D-PMD). For each matching pair of a video frame and procedural text, we generate a {success} example and various mistake examples by sampling alternate video frames: \textit{incomplete} execution, execution with the \textit{wrong verb} (e.g., wringing a cloth instead of folding), execution with the \textit{wrong noun} (e.g., folding paper instead of a cloth), and execution with both the \textit{wrong verb and noun} (e.g., opening a notepad instead of folding a cloth). Images cropped for space.}
    \label{fig:ego4d md}
\end{figure*}

\section{Problem Formulation and Dataset}
In this section, we define the extended problem of coherent PMD in an approachable manner for VLMs, describe how to apply VLMs to the problem, then lastly introduce a benchmark dataset we curated for evaluating coherent PMD.

\subsection{Defining Coherent PMD}\label{sec: explainable pmd}
The inputs for PMD are a short \textbf{procedural text} $T$ and a single \textbf{video frame} $F$, which may or may not visualize the successful completion of the procedure described in $T$.
Given these inputs, a system should return a binary \textbf{success decision} $y$ for whether the procedure has been successfully completed ($y=0$ indicates success, and $y=1$ indicates the detection of a mistake). In coherent PMD, it must additionally generate a \textbf{rationale} $R = (\mathcal{Q}, \mathcal{A})$, where $\mathcal{Q} = \{ Q_1, Q_2, \dots, Q_n \}$ and $\mathcal{A} = \{A_1, A_2, \dots, A_n \}$ are respectively sequences of $n$ yes-no questions and their predicted answers, which provide evidence for the decision.

\subsection{Applying VLMs to Coherent PMD}

As shown in Figure~\ref{fig: self dialog}, we elicit a rationale from VLMs through a self-generated visual dialog \cite{Das_2017_CVPR} consisting of \textbf{visual question generation} (VQG) based on the procedural text and dialog history, and \textbf{visual question answering} (VQA) based on the video frame. The evidence compiled in this rationale then conditions a \textbf{success classification} for whether the procedure has been completed in the given video frame. This structure goes beyond past approaches for PMD using VLMs; while \citet{pmlr-v232-du23b} only elicited success classification, \citet{bao-etal-2023-foundation} used procedure-agnostic prompts to caption images before classification, nonetheless disregarding this information in quantitative evaluations.

\paragraph{VQG.}
We prompt the VLM to generate a series of questions given the procedural text (and previous questions and answers in later iterations, enabling deductive reasoning). To encourage logical, diverse questions, we apply greedy beam search to generate 4 candidates, from which we select the most likely candidate not generated previously.

\paragraph{VQA.}
After a question is generated, it is answered by the VLM given only the question and video frame.
If the probability of the chosen answer (i.e., ``Yes'' or ``No'') exceeds an answer sureness threshold of 60\%, we append it to the dialog history, otherwise we append ``Unsure.''\footnote{Unsure answers are excluded from example-level informativeness (defined in Section~\ref{sec:pmd metrics}), and excluded from previous questions and answers in metric calculations.}

\paragraph{Success classification.}
After each iteration of VQG and VQA, we prompt the VLM to judge whether the procedure has been successfully executed based on the video frame and entire dialog history. The VLM's decision is made using a mistake confidence threshold $\tau$ (tuned on the validation data for each approach) on its mistake likelihood. The prompt and answer for this step are excluded from the dialog history in future iterations.

\paragraph{Stopping criteria.}\label{sec:stopping criteria}
To prevent over-generating evidence, which could introduce noise and degrade PMD accuracy, we implement an early stopping mechanism to determine whether to stop generating questions based on the success likelihood after each iteration. The self-dialog stops early (i.e., before a maximum number of questions $n^*$ are generated) if the success likelihood \textit{stabilizes} (i.e., changes by less than $\delta$ for two consecutive iterations) or becomes \textit{highly confident} (i.e., subceeds $\epsilon$ or exceeds $1-\epsilon$). $n^*$ is fixed at 10, while $\delta$ and $\epsilon$ are tuned based on the validation data for each presented approach.


\subsection{Constructing a Dataset for PMD}\label{sec:dataset}

We follow \citet{pmlr-v232-du23b} in recasting Ego4D \cite{Ego4D2022CVPR}, a large-scale egocentric video dataset for everyday activities with dense annotations for various aspects of the videos, into an offline mistake detection format, but expand the diversity of mistake types studied there.
Ego4D's hand and object interactions data subset includes videos of physical actions being performed with various objects. Each video is annotated with narrations describing fine-grained procedures being performed, each with timestamps for when it begins and ends, and category labels for the verb and noun characterizing the procedure. This makes an ideal testbed for evaluating VLMs' understanding of real-world actions in video frames, but the data is not formulated for PMD. We thus apply several preprocessing steps to the data to create a new Ego4D for Procedural Mistake Detection (Ego4D-PMD) benchmark that includes successful cases and a breadth of mistake types for each annotated procedure, visualized in Figure~\ref{fig:ego4d md}.

\begin{figure}
    \centering
    \includegraphics[width=0.91\linewidth]{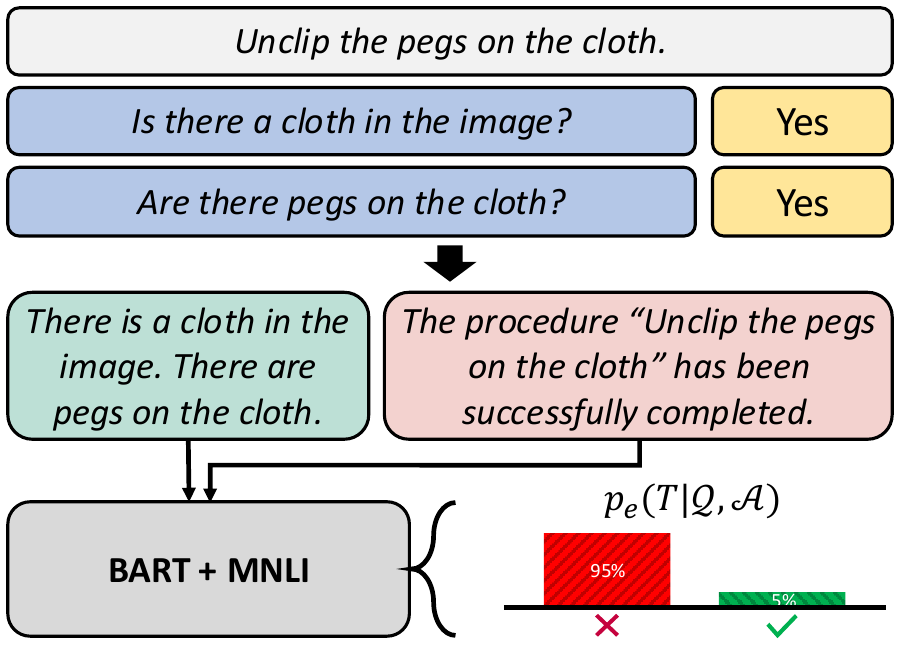}
    \caption{Using BART \cite{lewis-etal-2020-bart} fine-tuned on MNLI \cite{williamsBroadCoverageChallengeCorpus2017} to judge success.}
    \vspace{-15pt}
    \label{fig: nli model}
\end{figure}

Specifically, we form a successful case from each video clip in Ego4D by pairing its postcondition frame with its natural language narration of the procedure converted into imperative form. For each successful example, we generate several types of mistake examples paired with the source video's procedural text. To simulate the procedure being incomplete, we follow \citet{pmlr-v232-du23b} and sample another frame at the procedure's precondition time. To simulate an incorrect action being applied and/or the action being applied to an incorrect object or ingredient, we sample alternative video clips with mismatched verbs and nouns in their narration texts, in line with \citet{li2025mistakeattributionfinegrainedmistake}. Additional details about these preprocessing steps and summary statistics for the Ego4D-PMD dataset are presented in Appendix~\ref{apx: dataset details}. To conserve compute, we randomly sample a subset of 10,000, 500, and 2,000 examples respectively from the training, validation, and testing partitions (evenly split between success and mistake cases) for the forthcoming experiments.

%% file: latex/sections/3-methods.tex
\section{Evaluating Coherence in PMD}\label{sec:pmd metrics}
Next, we describe our application of a fine-tuned NLI model to calculate two evaluation metrics for coherent PMD: \textbf{relevance} of questions and \textbf{informativeness} of answers to those questions.

\subsection{Using NLI Models to Judge Success}\label{sec: pmd nli}
As shown in Figure~\ref{fig: nli model}, LMs fine-tuned for NLI can estimate the sufficiency of visual questions and answers in rationalizing whether a procedure was successfully completed. This requires an {NLI model} $f_e$ (which returns a probability that an input premise string entails a hypothesis string), a {premise transformation} $t_{p}$ (which converts a question $Q$ and answer $A$ into a declarative statement to add to the premise), and a {hypothesis prompt template} $t_h$ (which creates a hypothesis about the success of the procedure in $T$).
We then calculate the NLI model's probability for the success of procedure $T$ based on the rationale $R = (\mathcal{Q}, \mathcal{A})$:

\vspace{-12pt}

\begin{equation*}
     p_{e}(T|\mathcal{Q}, \mathcal{A}) = f_e\bigl( t_h(T) \vert \{ t_{p}(Q_i, A_i) \vert 1 \leq i \leq n \} \bigr)
\end{equation*}

We implement $f_e$ with BART \cite{lewis-etal-2020-bart} fine-tuned on the large-scale MultiNLI dataset \cite{williamsBroadCoverageChallengeCorpus2017},\footnote{See \url{https://huggingface.co/facebook/bart-large-mnli}.} applying softmax over its logits for entailment and contradiction to get an entailment probability. We follow \citet{srinivasan2024selective} and prompt a foundational LM to implement $t_{p}$.\footnote{To conserve GPU memory, we later use the evaluated VLM's LM backbone for rephrasing. Details in Appendix~\ref{apx: pmd rephrase}.} 
For the procedural text $T$, we choose a success prompt template $t_{h}$ ``The procedure $\langle P \rangle$ has been successfully executed.''\footnote{A template was used as we found complex procedural texts were unlikely to be rephrased accurately, degrading correlation with human judgments (see Appendix~\ref{apx: pmd human study details}).}

\subsection{Relevance}\label{sec:relevance}

A coherent decision in PMD should be supported by relevant questions about the state of the environment.\footnote{For example, given a procedure ``In a bowl, add the cut cherry tomatoes'' \cite{peddi2023captaincook4d}, the question ``Are there tomatoes in the bowl?'' is relevant to the success of the procedure, while the question ``Is the bowl blue?'' is not.} We measure the \textbf{relevance} of a question $Q'$ to the success of a procedure $T$, given previous questions $\mathcal{Q}$ and their answers $\mathcal{A}$, as follows:

\vspace{-12pt}
\begin{equation*}
\begin{split}
        \texttt{Rel}(Q'\vert T, \mathcal{Q}, \mathcal{A}) = &  \bigl| p_{e}(T|\mathcal{Q} \cup Q', \mathcal{A} \cup \text{``{No}''}) \\
        & - p_{e}(T|\mathcal{Q} \cup Q', \mathcal{A} \cup \text{``{Yes}''}) \bigr|
\end{split}
\end{equation*}

This definition quantifies how much impact the answer to the proposed question $Q'$ can have on the success probability (as estimated by the NLI model). If the success probability is similar for ``Yes'' and ``No'' answers, this suggests that $Q'$ would not reveal pertinent information (i.e., beyond that in $\mathcal{Q}$ and $\mathcal{A}$) about whether the procedure in $T$ was successfully executed by the user, and thus the relevance would be low. If the success probabilities vary widely depending on the answer, this suggests that $Q'$ can reveal important new information to help make the decision.

\paragraph{Example-level relevance.} 
To reward systems that propose consistently relevant questions in our evaluations, we summarize the relevance of a sequence of questions generated for a particular example by taking the mean question relevance with respect to previous questions and answers:

\vspace{-12pt}

\begin{equation*}
  \frac{1}{n} \sum\limits_{i = 1}^{n} \texttt{Rel} \bigl( Q_i|T, \{ Q_j: j < i \}, \{ A_j: j < i \} \bigr)
\end{equation*}


\subsection{Informativeness}\label{sec:informativeness}

Since a relevant question does not guarantee an informative answer,\footnote{For example, in the procedure ``In a bowl, add the cut cherry tomatoes,'' ``Are there tomatoes in the bowl?'' is a relevant question, but a ``Yes'' answer is insufficient to confirm 100\% completion (more \textit{tomatoes} could be outside the bowl).} and VLM errors in answering questions could unintentionally introduce conflicting information, it is necessary to evaluate the sufficiency of VLMs' answers in justifying a PMD decision. To achieve this, we measure the \textbf{informativeness} of a predicted answer $A'$ for a question $Q'$ to the success of a procedure $T$ (given previous questions and answers $\mathcal{Q}$ and $\mathcal{A}$) as follows:

\vspace{-12pt}

{
\small

\begin{equation*}
    \texttt{Inf}(A'|Q', T, \mathcal{Q}, \mathcal{A}) = 1 - H\bigl( p_{e}(T|\mathcal{Q} \cup Q', \mathcal{A} \cup A') \bigr)
\end{equation*}
}

$H$ is the binary Shannon entropy of the success probability $p_{e}$, calculated by $H(p) = -p \log_2 p - (1 - p) \log_2 (1 - p), p \in \lbrack 0, 1 \rbrack$. 
This definition for informativeness quantifies how much information an answer to a question provides in determining the success of the procedure. As such, if the success probability given this answer $A'$ to $Q'$ is confident (i.e., very low or high), this indicates that $A'$ (along with $\mathcal{Q}$ and $\mathcal{A}$) are sufficient to make a decision, and thus informativeness is high. Conversely, a success probability close to 50\% suggests the evidence gathered thus far is insufficient, yielding low informativeness. 
Informativeness is expressed as a number of bits between 0 and 1.

\paragraph{Reference-adjusted informativeness.}
We also wish to account for cases where the evidence gathered in questions and answers indicates the wrong PMD decision. To do so, we define the NLI model's PMD belief $y_e(T|\mathcal{Q}, \mathcal{A})$ as 1 (mistake) if $p_e(T|\mathcal{Q}, \mathcal{A}) < 0.5$, else 0 (success). Given the ground truth PMD label $y^*$, we then define the \textbf{reference-adjusted informativeness} to be negative if the NLI model judges the evidence gathered as indicating the wrong success decision:

\vspace{-12pt}

{
\small
\begin{multline*}
\texttt{Inf}^*(A'|Q', T, \mathcal{Q}, \mathcal{A}, y^*) = \\
    \begin{cases}
        \texttt{Inf} \bigl( A'|Q', T, \mathcal{Q}, \mathcal{A} \bigr), & y_e (T|
        \mathcal{Q} \cup Q', \mathcal{A} \cup A') = y^* \\
        - \texttt{Inf} \bigl( A'|Q', T, \mathcal{Q}, \mathcal{A} \bigr), & \text{else}
    \end{cases}
\end{multline*}
}


\paragraph{Example-level informativeness.}
To summarize the sufficiency of information gathered throughout the self-dialog (which may have some uninformative answers), we take the maximum (reference-adjusted) informativeness across the self-dialog:


\vspace{-12pt}

{
\small
\begin{equation*}\label{eq: verifiability}
      \max\limits_{1 \leq i \leq n} \texttt{Inf$^*$}(A_i|Q_i, T, \{Q_j: j<i \}, \{A_j: j<i \}, y^*)
\end{equation*}

}

%% file: latex/sections/4-results.tex
\section{Rationale Coherence Interventions}
To validate these metrics and examine the relationship between rationale coherence, PMD accuracy, and rationale generation efficiency, we next introduce two inference-time interventions to encourage VLMs to select more coherent questions from generated candidates, as well as a preference optimization approach to encourage generating more coherent candidates based on coherence metrics.

\subsection{Coherent Question Selection}\label{sec:question selection}
While a typical beam search would use a sequence \textit{likelihood-based} approach to rank candidate questions, an alternative is to re-rank candidates using the reference-free coherence metrics introduced in Section~\ref{sec:pmd metrics}. This could encourage selecting questions that are likely to bring in new, salient, and helpful information.
Furthermore, as adaptive information-seeking is a core component of humans' reasoning capabilities \cite{coenen2019asking}, 
we will supplement candidate questions with candidates generated through in-context learning from human-written examples (which we assume are reasonably coherent and effective).


Next, we introduce these two approaches we use to augment the candidate question pool for more coherent candidates: \textit{coherence-based re-ranking} and \textit{candidate generation through in-context learning}. We then compare their performance on Ego4D-PMD with that of vanilla VLMs.

\subsubsection{Coherence-Based Question Selection}\label{sec:coherence ranking}
We implement a coherence-based candidate question re-ranking approach as follows. Given a set of question candidates $\hat{\mathcal{Q}}$ for procedural text $T$ along with previous confidence-filtered questions $\mathcal{Q}$ and answers $\mathcal{A}$, we can select the best question $Q^*$ by maximizing the product of relevance and potential informativeness across all $Q\in\hat{Q}$:

\vspace{-12pt}

{
\begin{equation*}
    \texttt{Rel}(Q|T, \mathcal{Q}, \mathcal{A}) 
    \mathop{\times \:\: \text{max} \:\: \texttt{Inf}}_{A \in \{\text{\textit{Yes}}, \text{\textit{No}}\}}(A|Q,T, \mathcal{Q}, \mathcal{A})
\end{equation*}
}

This ranking prioritizes well-rounded questions that can yield impactful information for success classification which leads to the most confidence. $Q^*$ is then concatenated to the dialog history and answered by the VLM as described earlier.

\subsubsection{In-Context Learning Augmentation}\label{sec:pmd icl details}
Applying in-context learning in coherent PMD is not straightforward, as it would require reasoning over multiple images and dialogs of varying length.
Instead, we use in-context learning to improve the text-based VQG step by providing examples of human-written questions.
We achieve this by annotating 20 procedures from the Ego4D training data with 3 reasonable questions one could ask about a given procedure to judge its success.\footnote{All questions and more details in Appendix~\ref{apx: pmd icl}.} We prompt the VLM with these example procedures and questions, the current procedure at hand, and the previous 2 questions proposed by the VLM (as available) to incorporate information the VLM already collected. We then generate 4 additional candidate questions using the same constraints described earlier. To minimize the impact of ordering, in-context examples are randomly shuffled in every prompt.\footnote{Examples are equivalently shuffled in compared results.}


\subsection{Coherent Question Generation}\label{sec:coherence training}
While the above approaches may boost the coherence of PMD rationales, they have limitations. First, since they are training-free, they do not improve the internal coherence of VLMs, rather they only filter and augment their outputs. Further, these interventions take significantly more time and compute due to the need to evaluate the coherence of candidate questions (which requires prompting an NLI model) and generate questions twice (once based on the self-dialog history and once with in-context learning). In practical applications like task guidance, it may be advantageous to use specialized VLMs to improve reliability and speed.\footnote{To explore this further, we provide a runtime analysis of various configurations of VLM self-dialog in Appendix~\ref{apx:timing analysis}.} As such, we also explore whether VLMs can be fine-tuned to generate more coherent questions using our automated coherence metrics.

Specifically, we apply direct preference optimization (DPO; \citealp{rafailov2023direct}) with quantized low-rank adaptation (QLoRA; \citealp{hu2022lora,dettmers2024qlora}) to fine-tune a VQG adapter for LLaVA.
We generate training data by first running inference over the Ego4D-PMD training data with coherence-based candidate question re-ranking and additional candidates generated through in-context learning.\footnote{Ablation without in-context learning in Appendix~\ref{apx:dpo ablation}.} We then fine-tune VLMs on pairs of chosen and rejected candidate questions for each self-dialog turn based on their coherence ranking score.\footnote{More details and validation results in Appendix~\ref{apx:dpo details}.} The top ranked question is always chosen, while a rejected question is sampled from the bottom half of candidates.
At inference time, the trained adapter is applied during VQG, and disabled for other steps.

\section{Experimental Results}
In this section, we evaluate VLMs on coherent PMD off-the-shelf and under the previously introduced interventions.
Metrics include mistake detection accuracy, mean example relevance and reference-adjusted informativeness of rationales (defined in Section~\ref{sec:pmd metrics}). We add two metrics for self-dialog efficiency: the average \textbf{number of iterations} that occurred before stopping (i.e., the length of the dialog), and the average \textbf{information gain} in the success likelihood across all iterations in bits (i.e., how much information the VLM got from it).
We lastly show that our proposed metrics visualize common behaviors of VLMs, enabling a panoptic understanding of their performance.

\paragraph{Evaluated models.}

We specifically evaluate InstructBLIP \cite{instructblip}, LLaVA 1.5-7B \cite{liu2023llava}, and Llama 3.2-Vision-11B \cite{dubey2024llama3herdmodels}, small open-source VLMs feasible for online use (important for real-world applications like task guidance).
They apply different architectures and training strategies to integrate vision into their LMs.
LLaVA and InstructBLIP were not trained on Ego4D, while Llama 3's training data was not disclosed.
For a reference point with state-of-the-art proprietary VLMs, we additionally include results with off-the-shelf GPT-4o~\cite{openai2024gpt4ocard}.
Additional details and prompt templates can be found in Appendix~\ref{apx: pmd prompt details} and \ref{apx:gpt results}.\footnote{In Appendix~\ref{apx:exp free results}, we include an additional evaluation of VLMs without generating rationales (as done in prior work).}

\paragraph{Human accuracy.}

To create a reference point for VLMs' PMD accuracy and a proxy for data quality and objectiveness, we recruited human annotators for PMD classification. 100 random testing examples were labeled by 3 annotators each, yielding a majority-vote human accuracy of 72.0\%. This suggests that PMD itself is fairly subjective and difficult for humans, further necessitating coherent rationalization. More details about this annotation are provided in Appendix~\ref{apx:human accuracy annotation}.

\begin{table}[!t]
    \centering
    \setlength{\tabcolsep}{3.5pt}




    
    \normalsize
    \textbf{InstructBLIP}

    \vspace{1pt}

    \footnotesize
    \begin{tabular}{ccccccc}\toprule
        \textbf{Rank} & \textbf{ICL} & \textbf{Acc.} $\uparrow$ & \textbf{Rel.} $\uparrow$ & \textbf{Inf.} $\uparrow$ & \textbf{\# Iter.} $\downarrow$ & \textbf{I. Gain} $\uparrow$ \\\cmidrule(lr){1-2}\cmidrule(lr){3-3}\cmidrule(lr){4-5}\cmidrule(lr){6-7}
        L & \xmark & 63.5 & 17.5 & .224 & \textbf{2.84} & .263 \\
        L & \cmark & 65.2 & 13.9 & .340 & 4.71 & .358 \\
        C & \xmark & 64.6 & 25.5 & .281 & 3.46 & .293 \\  
        C & \cmark & \textbf{66.6} & \textbf{35.2} & \textbf{.359} & 3.47 & \textbf{.359} \\\bottomrule
    \end{tabular}

    \vspace{6pt}

    \normalsize
    \textbf{LLaVA}

    \vspace{1pt}

    \footnotesize
    \begin{tabular}{ccccccc}\toprule
        \textbf{Rank} & \textbf{ICL} & \textbf{Acc.} $\uparrow$ & \textbf{Rel.} $\uparrow$ & \textbf{Inf.} $\uparrow$ & \textbf{\# Iter.} $\downarrow$ & \textbf{I. Gain} $\uparrow$ \\\cmidrule(lr){1-2}\cmidrule(lr){3-3}\cmidrule(lr){4-5}\cmidrule(lr){6-7}        
        L & \xmark & 60.7 & 40.3 & .259 & 3.25 & .435 \\
        L & \cmark & 61.8 & 36.5 & .272 & 3.34 & .429 \\
        C & \xmark & 61.4 & 66.5 & .321 & \textbf{3.06} & .540 \\
        C & \cmark & \textbf{67.8} & \textbf{75.5} & \textbf{.464} & 3.46 & \textbf{.663} \\\bottomrule
    \end{tabular}

    \vspace{6pt}

    \normalsize
    \textbf{Llama 3}

    \vspace{1pt}

    \footnotesize
    \begin{tabular}{ccccccc}\toprule
        \textbf{Rank} & \textbf{ICL} & \textbf{Acc.} $\uparrow$ & \textbf{Rel.} $\uparrow$ & \textbf{Inf.} $\uparrow$ & \textbf{\# Iter.} $\downarrow$ & \textbf{I. Gain} $\uparrow$ \\\cmidrule(lr){1-2}\cmidrule(lr){3-3}\cmidrule(lr){4-5}\cmidrule(lr){6-7}        
        L & \xmark & 61.0 & 16.5 & .275 & 4.70 & .223 \\
        L & \cmark & 59.1 & 15.9 & .317 & 6.51 & .256 \\
        C & \xmark & 60.2 & 25.2 & .341 & 6.35 & .264 \\
        C & \cmark & \textbf{61.7} & \textbf{52.5} & \textbf{.436} & \textbf{3.59} & \textbf{.379} \\\bottomrule
    \end{tabular}

    \vspace{6pt}
    
    \normalsize
    \textbf{GPT-4o}

    \vspace{1pt}

    \footnotesize
    \begin{tabular}{ccccccc}\toprule
        \textbf{Rank} & \textbf{ICL} & \textbf{Acc.} $\uparrow$ & \textbf{Rel.} $\uparrow$ & \textbf{Inf.} $\uparrow$ & \textbf{\# Iter.} $\downarrow$ & \textbf{I. Gain} $\uparrow$ \\\cmidrule(lr){1-2}\cmidrule(lr){3-3}\cmidrule(lr){4-5}\cmidrule(lr){6-7}
    L & \xmark & 55.4 & 54.0 & .220 & 1.84 & .793 \\\bottomrule
    \end{tabular}

    \normalsize
    \vspace{-3pt}
    \caption{Ego4D-PMD test set results for GPT-4o \cite{openai2024gpt4ocard} and likelihood-based (L) and coherence-based (C) candidate question ranking approaches, with optional supplementary candidates generated through in-context learning (ICL).}

    
    \label{tab:selection metrics}

    \centering
    \setlength{\tabcolsep}{3.5pt}




    \vspace{4pt}

    \normalsize
    \textbf{LLaVA + DPO}

    \vspace{1pt}

    \footnotesize
    \begin{tabular}{ccccccc}\toprule
        \textbf{Rank} & \textbf{ICL} & \textbf{Acc.} $\uparrow$ & \textbf{Rel.} $\uparrow$ & \textbf{Inf.} $\uparrow$ & \textbf{\# Iter.} $\downarrow$ & \textbf{I. Gain} $\uparrow$ \\\cmidrule(lr){1-2}\cmidrule(lr){3-3}\cmidrule(lr){4-5}\cmidrule(lr){6-7}        
        L & \xmark & 62.2 & 75.7 & .318 & 2.33 & .617 \\
        L & \cmark & 63.7 & 58.5 & .330 & 2.67 & .548 \\
        C & \xmark & 62.3 & 92.2 & \textbf{.340} & 2.06 & .719 \\
        C & \cmark & \textbf{64.2} & \textbf{95.0} & .304 & \textbf{1.81} & \textbf{.742} \\\bottomrule
    \end{tabular}
    \normalsize

    \vspace{-3pt}
    \caption{Ego4D-PMD test set results for LLaVA with coherence-based fine-tuning through DPO.}

    
    \label{tab:generation metrics}
\end{table}

\subsection{Coherent Question Selection Results}
Experimental results for coherent question selection interventions are presented in Table~\ref{tab:selection metrics}.\footnote{Hyperparameters and validation results in Appendix~\ref{apx:selection hyperparam}.}
We find that \textit{coherence-based re-ranking and in-context learning}\footnote{We visualize the distribution of question sources (i.e., in-context learning vs. full self-dialog history) in Appendix~\ref{apx:icl hist}.} \textit{both sharply improve the relevance and informativeness} in all models, reaching a respective 75.5\% and 0.464 bits in LLaVA. This suggests that \textit{questions that VLMs find most likely are not naturally the most coherent}. Interestingly, accuracy also jumps sharply for InstructBLIP and LLaVA to a maximum of 67.8\%, demonstrating that \textit{coherent rationales are valuable to accurate PMD}, although accuracy still lags slightly behind humans (72.0\%).
Information gain under these interventions is consistently higher, reaching a maximum of 0.663 bits in LLaVA. This suggests that \textit{VLMs can make more confident conclusions given more coherent rationales}.\footnote{We contextualize these results with a na{\"i}ve semantic diversity-based ranking baseline in Appendix~\ref{apx:distance based}.} Meanwhile, our best configurations outperform GPT-4o in accuracy, relevance, and informativeness at less than 10\% of its size \cite{abacha2025medecbenchmarkmedicalerror}. While GPT-4o takes fewer iterations and exhibits higher information gain (thus making faster and more confident decisions), its rigidity as a closed proprietary VLM prevents further improvements.

\subsection{Coherent Question Generation Results}\label{sec: dpo results}
When fine-tuned for coherent question generation,
Table~\ref{tab:generation metrics} shows that relevance drastically increases from a previous maximum of 75.5\% to 95.0\%.
Remarkably, this demonstrates that \textit{VLMs can learn to ask more coherent questions for PMD}, a task based on properties of the physical world.

However, we find that accuracy and informativeness drop slightly from the best results in Table~\ref{tab:selection metrics} to a respective maximum of 64.2\% and 0.340 bits. This suggests that \textit{asking more relevant questions is not enough to improve performance globally}. While more coherent rationales previously improved accuracy, this instead demonstrates a nontrivial relationship between coherence metrics and task accuracy. We suspect that asking highly relevant questions (which strongly indicate success if answered one way, otherwise a mistake) introduces a trade-off with question difficulty. For elaborate procedures, highly relevant questions may cover multiple aspects or states of a scene. As such, answering these more complex questions may be difficult for VLMs.
For example, while the VLM-generated question ``Is the soil placed around the seedling with the trowel in the person's hand?'' covers the success conditions of the procedure ``Put some soil around the tomato seedling with the gardening trowel in your hand,'' answering it requires understanding spatial relations between several objects (i.e., \textit{soil}, \textit{seedling}, \textit{trowel}, and \textit{hand}). Future work may explore methods to encourage generating simpler questions.\footnote{As an initial inquiry here, we experiment with applying an exponential length penalty to VQG in Appendix~\ref{apx:dpo ablation 2}.}

Meanwhile, we find that fine-tuning reduces the average number of iterations taken to come to a conclusion by about 1.25, reaching a minimum of 1.81 (matching that of GPT-4o). This is especially important for online PMD, which requires a fast reaction.
Further, the average information gain reaches a new peak of 0.742, nearly matching that of GPT-4o. This suggests that \textit{coherence-based fine-tuning can empower VLMs to make more confident decisions faster}. 
We lastly observe that coherence-based ranking and in-context learning have a smaller impact on performance metrics
in fine-tuned LLaVA than in the base model (Table~\ref{tab:selection metrics}). This suggests that \textit{fine-tuning VLMs using our coherence metrics reduces the need for inference-time interventions for coherence}.

\begin{figure*}
    \centering
    
    \footnotesize

    \setlength{\columnsep}{0pt}
    
    \begin{multicols}{4}
    
    \textbf{Vanilla LLaVA}
    
    \vspace{3pt}
    
    \includegraphics[width=1.0\linewidth,trim={4em 2em 0 5em},clip]{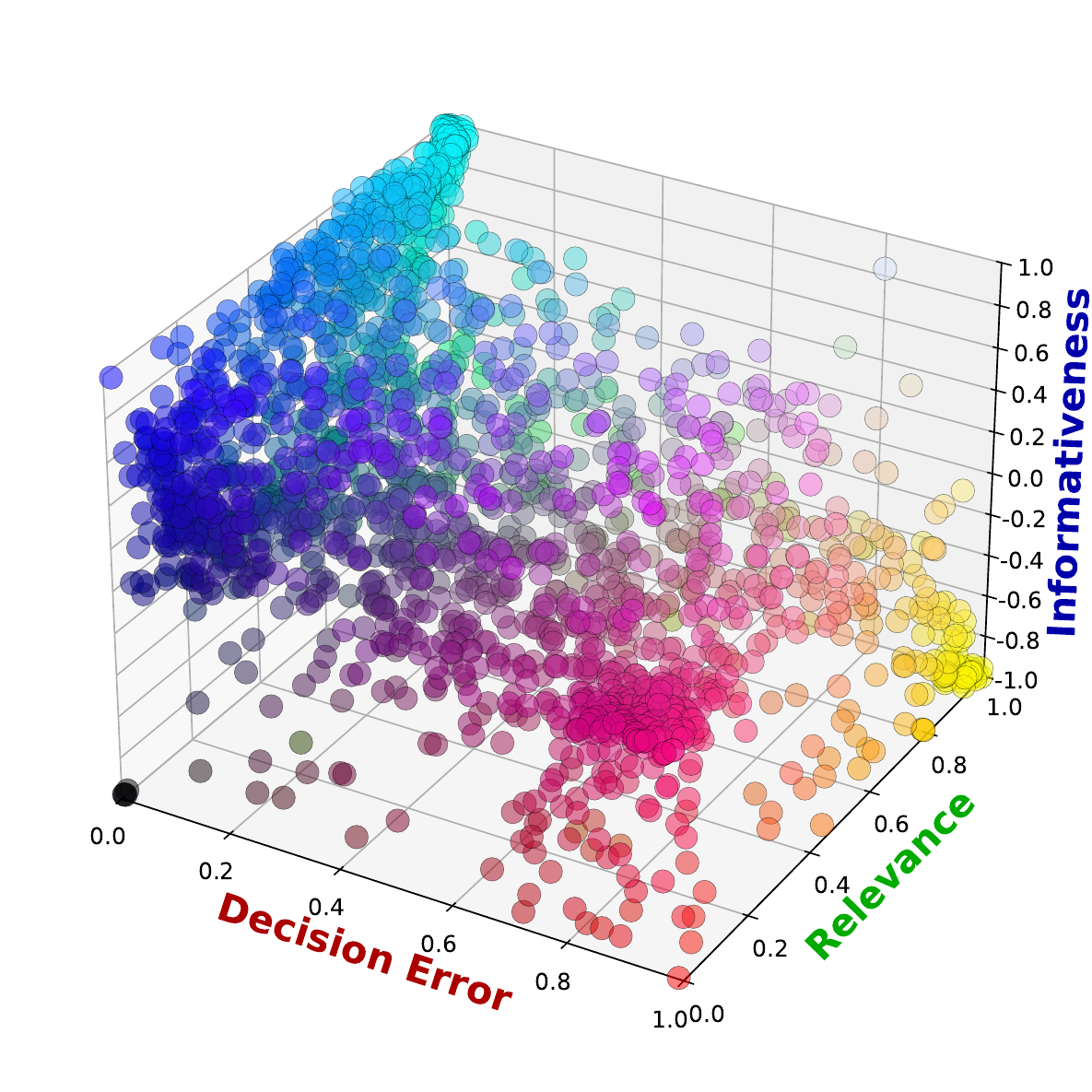}

    \vspace{3pt}

    \textbf{+ Coherence-Based Ranking}

    \vspace{3pt}
    
    \includegraphics[width=1.0\linewidth,trim={4em 2em 0 5em},clip]{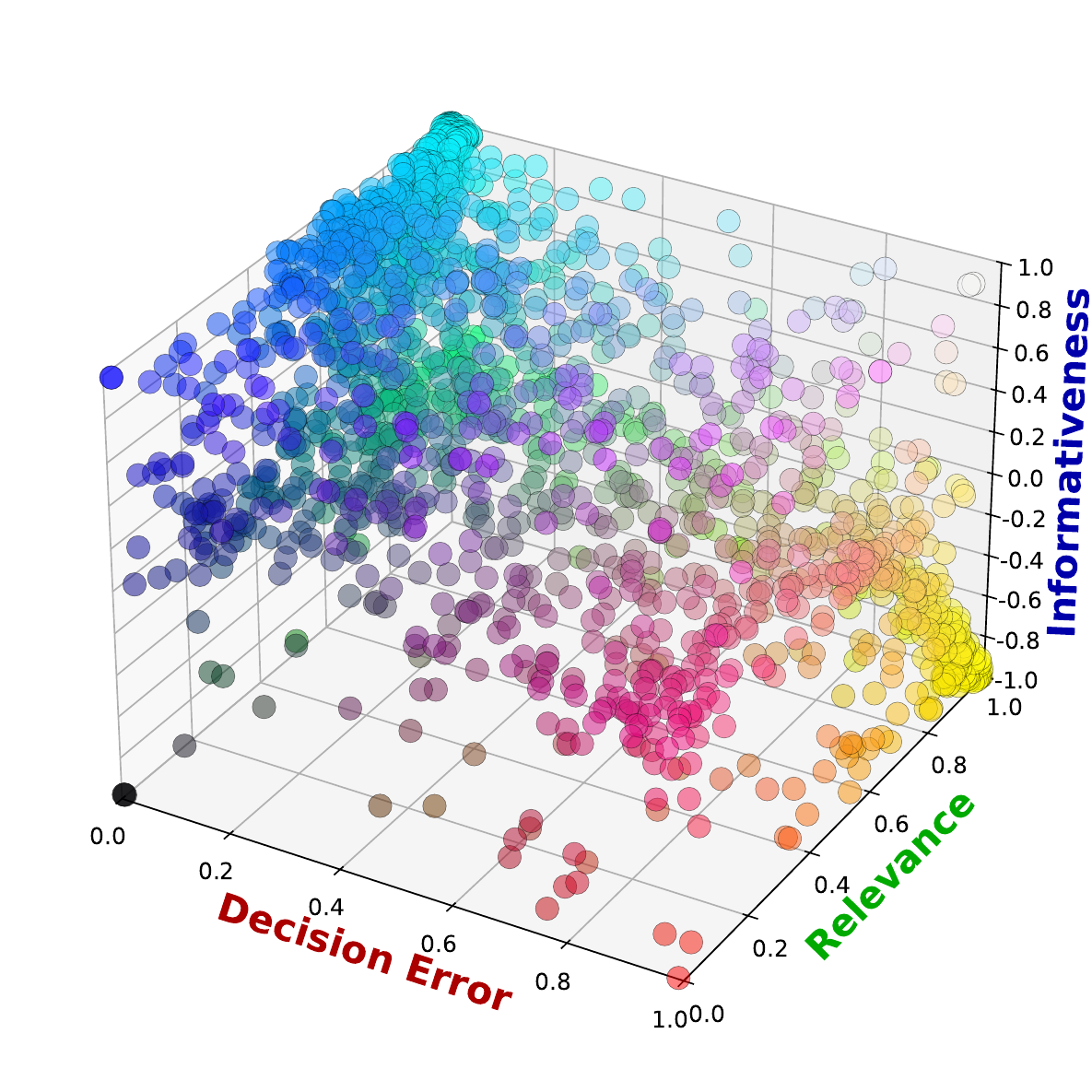}
    
    \vspace{3pt}

    \textbf{+ In-Context Learning}
    
    \vspace{3pt}
    
    \includegraphics[width=1.0\linewidth,trim={4em 2em 0 5em},clip]{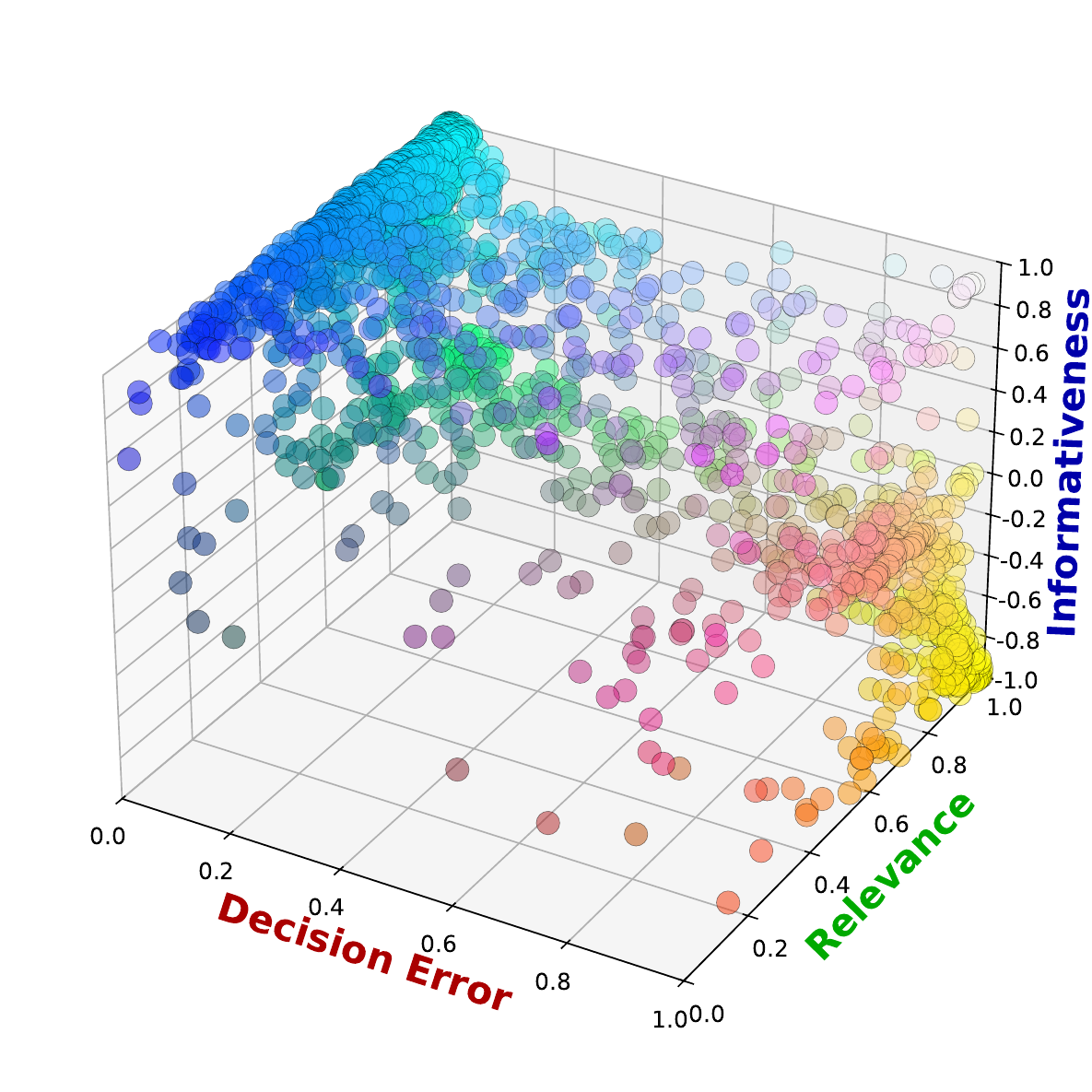}



    
    
    \vspace{3pt}

    \textbf{+ Coherence-Based DPO}

    \vspace{3pt}
    
    \includegraphics[width=1.0\linewidth,trim={4em 2em 0 5em},clip]{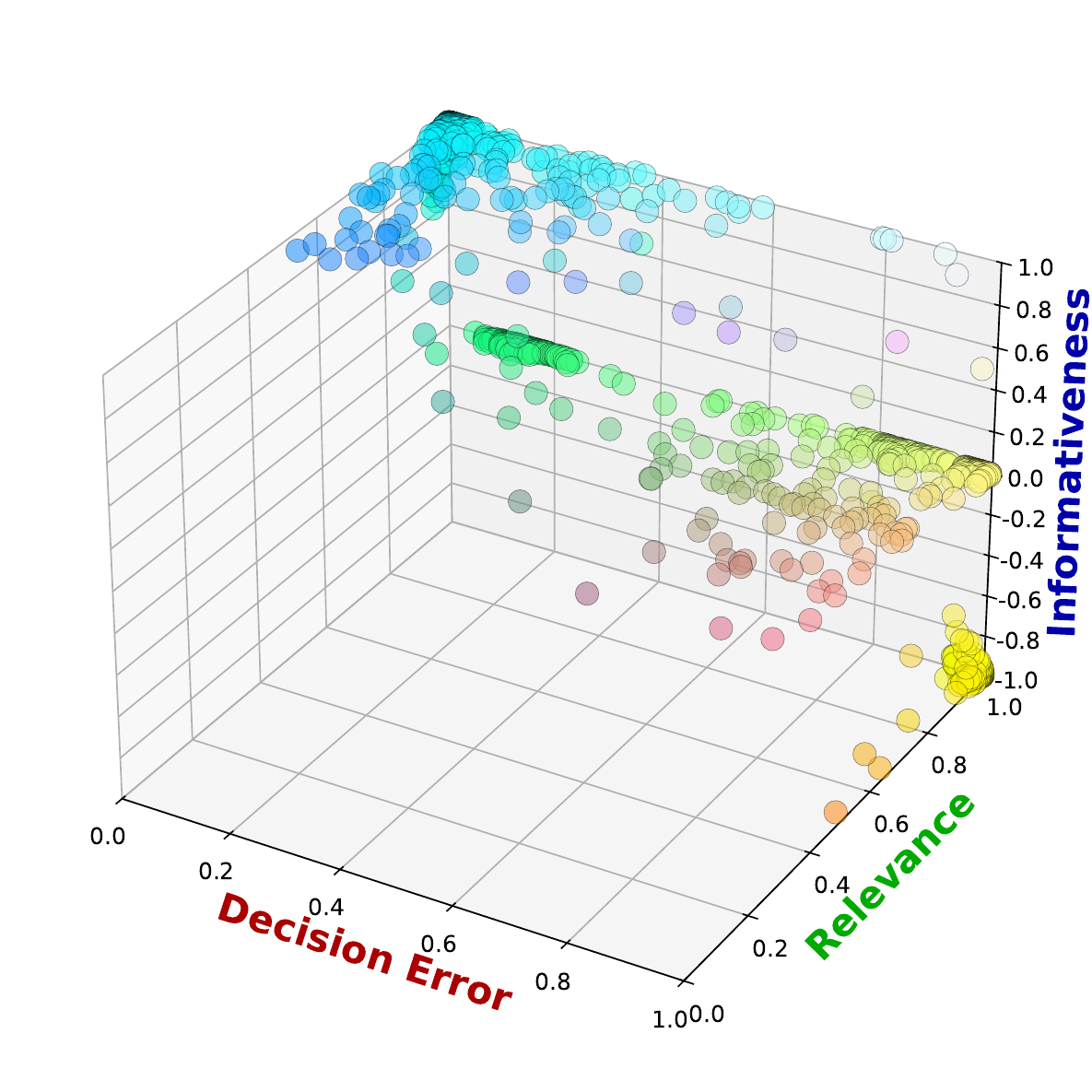}
    
    \end{multicols}
    
    \vspace{-23pt}
    
    \caption{Visualization of decision error, relevance, and reference-adjusted informativeness for configurations of LLaVA on Ego4D-PMD testing examples. Each data point's color indicates its position along each axis.}

    \vspace{-7pt}

\label{fig:cubes}
\end{figure*}

\subsection{Visualizing PMD Performance}\label{sec: final analysis}
An advantage of our automated coherence metrics is the ability to audit VLMs' reasoning behaviors. In Figure~\ref{fig:cubes}, we visualize the distribution of decision error, example-level relevance, and example-level informativeness of four representative approaches. Specifically, we compare vanilla LLaVA (i.e., with likelihood-based question ranking) to variants successively equipped with coherence-based question ranking, in-context learning, and coherence-based DPO fine-tuning.
For each example, decision error is calculated by how far the VLM's success likelihood is from being 100\% confident in the correct success label. 

Point colors indicate combinations of decision error, relevance, and informativeness, highlighting common local outcomes. Several examples are shown in Figure~\ref{fig:cube graph examples}. While cyan points indicate correct decisions with coherent rationales, red points indicate incorrect decisions with incoherent rationales, e.g., asking irrelevant questions about whether a person was wearing \textit{protective gear} in determining whether a \textit{screw} had been tightened. Black and indigo points indicate correct decisions with incoherent rationales, while white points indicate incorrect decisions with coherent rationales, suggesting inconsistencies in rationales or failures to interpret them (e.g., correctly identifying a \textit{bottle of mustard} on the \textit{countertop}, but then mistakenly detecting it on the \textit{floor}). Interestingly, green and yellow points with low informativeness typically indicate failures in VQA, e.g., unsure answers, missing the appearance of a \textit{trowel}, and hallucinating the appearance of a \textit{bottle}. These examples elucidate the nontrivial relationship between rationale coherence and PMD accuracy observed in earlier results; a variety of errors in both generating and interpreting rationales can cause downstream errors in PMD. We provide an extended discussion of these cases in Appendix~\ref{apx:cube graph examples}.

In comparing the plots globally, we see a virtual elimination of red, black, and indigo points with irrelevant and uninformative rationales. We also see a significant reduction in the range of informativeness under coherence-based fine-tuning. This suggests that when questions are answered correctly under this approach, it is highly informative to the success decision, and vice versa. This is in line with our earlier observation that coherence-based fine-tuning may encourage the generation of overly relevant and thus complex questions, which can make or break the rationale. There are also a large number of points with zero informativeness, i.e., unsure answers to these complex questions.


%% file: latex/sections/5-related-work.tex
\begin{figure*}
    \centering
    \scriptsize

    \scalebox{1.0}{
    \begin{minipage}{1.0\textwidth}
    \fboxsep0pt
    \setlength\columnsep{5pt}
    \begin{multicols}{4}

        \begin{examplebox}{A}{Pick up a sink brush from the kitchen slab.}{pcyan}

        \includegraphics[width=1\linewidth,trim={50px 60px 130px 140px},clip]{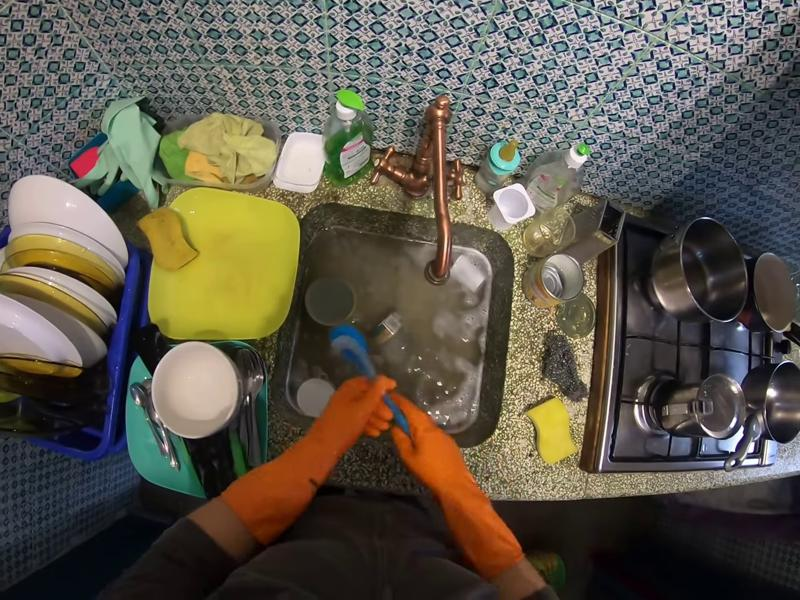} 

        {

        \centering
        \textbf{Label}: \textcolor{darkgreen}{\cmark} \hspace{20pt} 
        \textbf{Predicted}: \textcolor{darkgreen}{\cmark}
        
        }

        \vspace{2pt}

        \textbf{Rationale:} \textit{Is the sink brush...}

        
        \begin{enumerate}[noitemsep,leftmargin=8pt,topsep=0pt,labelsep=2pt]
            \item \textit{in the person's hand?} \texttt{Yes}
        \end{enumerate}
        
        \end{examplebox}
%

        \begin{examplebox}{E}{Fold the cut piece of cloth.}{pblue}

        \includegraphics[width=1\linewidth,trim={60px 60px 60px 90px},clip]{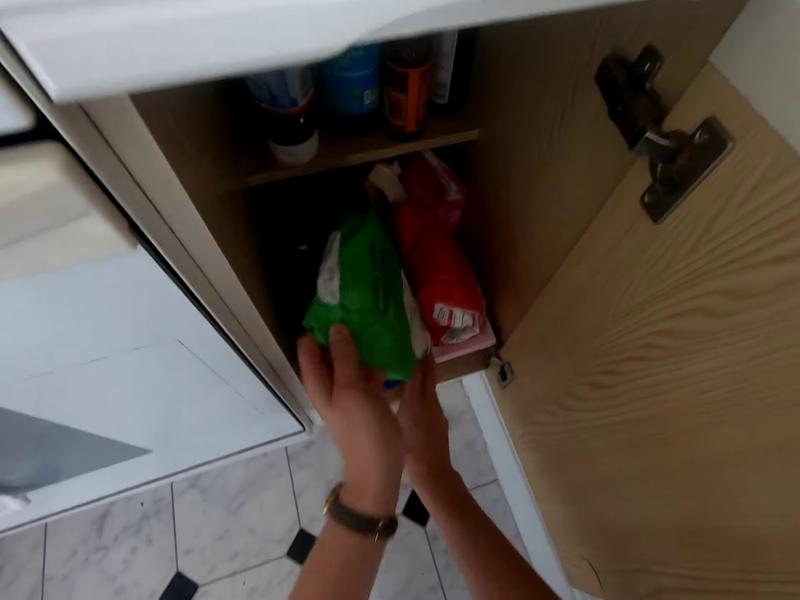} 

        {
        \centering
        
        \textbf{Label}: \textcolor{darkred}{\xmark} \hspace{20pt} 
        \textbf{Predicted}: \textcolor{darkred}{\xmark}
        
        }

        \vspace{2pt}

        \textbf{Rationale:} \textit{Is the person...}

        
        \begin{enumerate}[noitemsep,leftmargin=8pt,topsep=0pt,labelsep=2pt]
            \item \textit{working on a piece of cloth?} \texttt{No}
        \end{enumerate}
        
        \end{examplebox}

        \vspace{2pt}

%
        \begin{examplebox}{B}{Tighten the screw.}{pred}

        \includegraphics[width=1\linewidth,trim={100px 230px 100px 40px},clip]{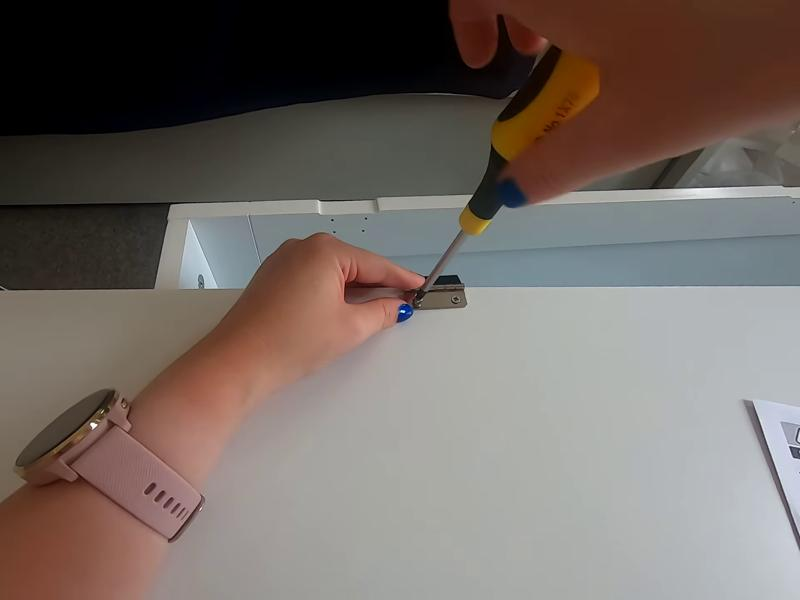} 

        {
        \centering
        
        \textbf{Label}: \textcolor{darkgreen}{\cmark} \hspace{20pt} 
        \textbf{Predicted}: \textcolor{darkred}{\xmark}
        
        }

        \vspace{2pt}

        \textbf{Rationale:} \textit{Is the person...}

        
        \begin{enumerate}[noitemsep,leftmargin=8pt,topsep=0pt,labelsep=2pt]
            \item \textit{wearing gloves?} \texttt{No}
            \item \textit{wearing protective gear?} \texttt{No}
            \item \textit{wearing a mask?} \texttt{No}
        \end{enumerate}
        
        \end{examplebox}

                \begin{examplebox}{F}{Put the bolt remover in the lawn tractor.}{pgreen}

        \includegraphics[width=1\linewidth,trim={80px 30px 80px 170px},clip]{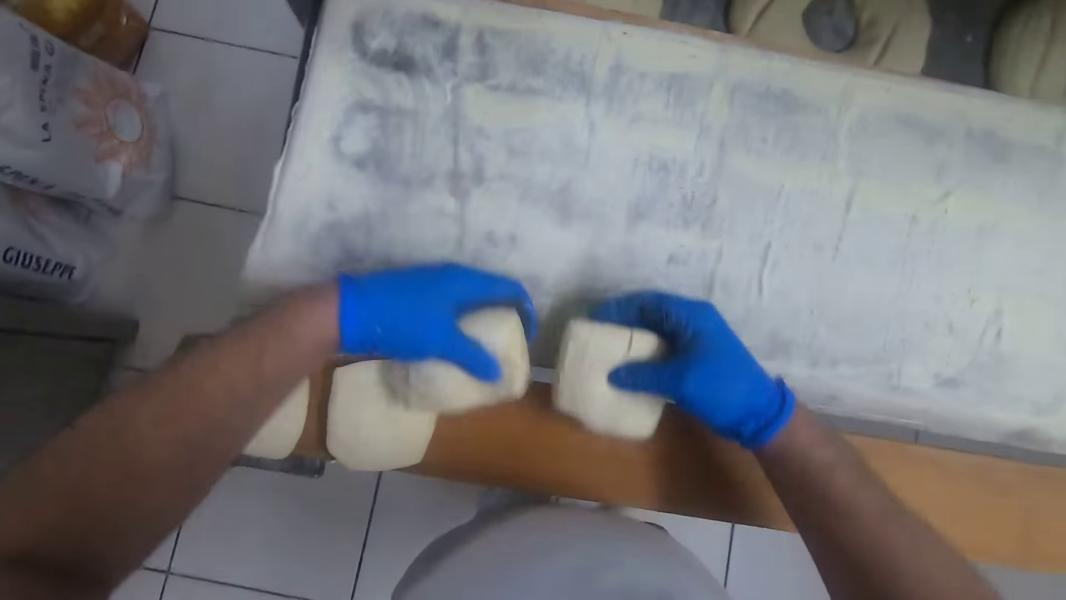} 

        {
        \centering
        
        \textbf{Label}: \textcolor{darkred}{\xmark} \hspace{20pt} 
        \textbf{Predicted}: \textcolor{darkred}{\xmark}
        
        }

        \vspace{2pt}

        \textbf{Rationale:} \textit{Is the bolt remover...}

        
        \begin{enumerate}[noitemsep,leftmargin=8pt,topsep=0pt,labelsep=2pt]
            \item \textit{in the lawn tractor?} \texttt{Unsure}
            \item \textit{inside the lawn tractor?} \texttt{Unsure}
            \item \textit{attached to the lawn tractor?} \texttt{Unsure}
        \end{enumerate}
        
        \end{examplebox}

        \vspace{2pt}

        \begin{examplebox}{C}{Paint the stone.}{pblack}

        \includegraphics[width=1\linewidth,trim={170px 90px 50px 100px},clip]
        {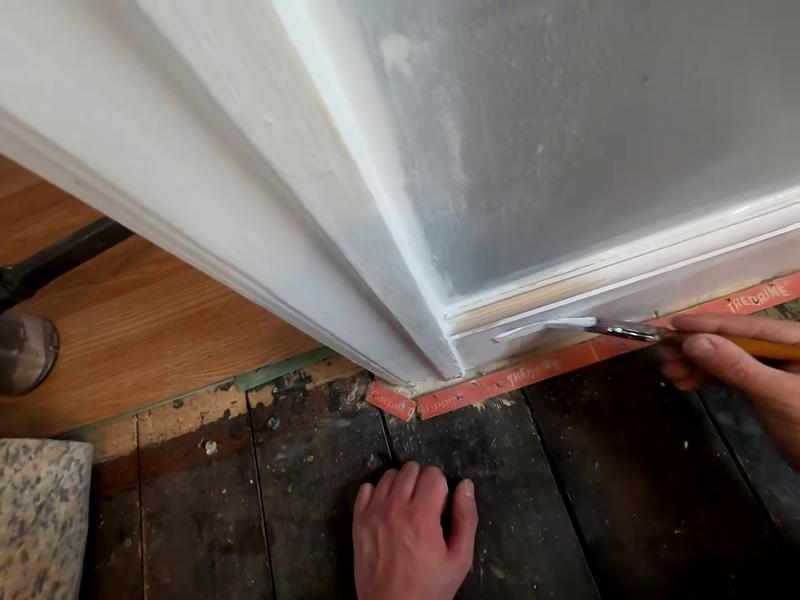} 

        {
        \centering
        
        \textbf{Label}: \textcolor{darkred}{\xmark} \hspace{20pt} 
        \textbf{Predicted}: \textcolor{darkred}{\xmark}
        
        }

        \vspace{2pt}

        \textbf{Rationale:} \textit{Is the person...}

        
        \begin{enumerate}[noitemsep,leftmargin=8pt,topsep=0pt,labelsep=2pt]
            \item \textit{wearing a shirt?} \texttt{Unsure}
        \end{enumerate}
        
        \end{examplebox}


        \begin{examplebox}{G}{Put the trowel in a bin.}{pyellow}

        \includegraphics[width=1\linewidth,trim={220px 180px 20px 2px},clip]{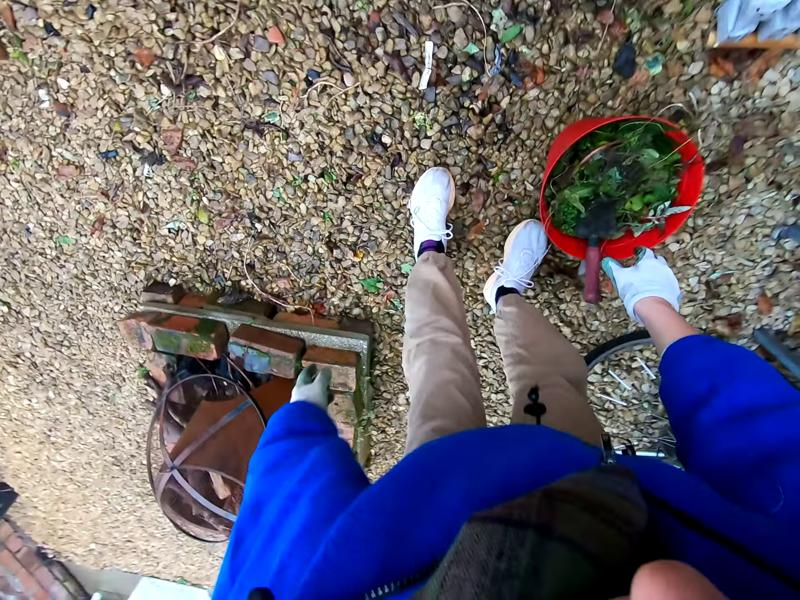} 

        {
        \centering
        
        \textbf{Label}: \textcolor{darkgreen}{\cmark} \hspace{20pt} 
        \textbf{Predicted}: \textcolor{darkred}{\xmark}
        
        }

        \vspace{2pt}

        \textbf{Rationale:} \textit{Is the trowel...}

        
        \begin{enumerate}[noitemsep,leftmargin=8pt,topsep=0pt,labelsep=2pt]
            \item \textit{in a bin?} \texttt{No}
        \end{enumerate}
        
        \end{examplebox}

        \begin{examplebox}{D}{Drop the bottle of mustard on the countertop.}{pwhite}

        \includegraphics[width=1\linewidth,trim={80px 165px 0px 40px},clip]{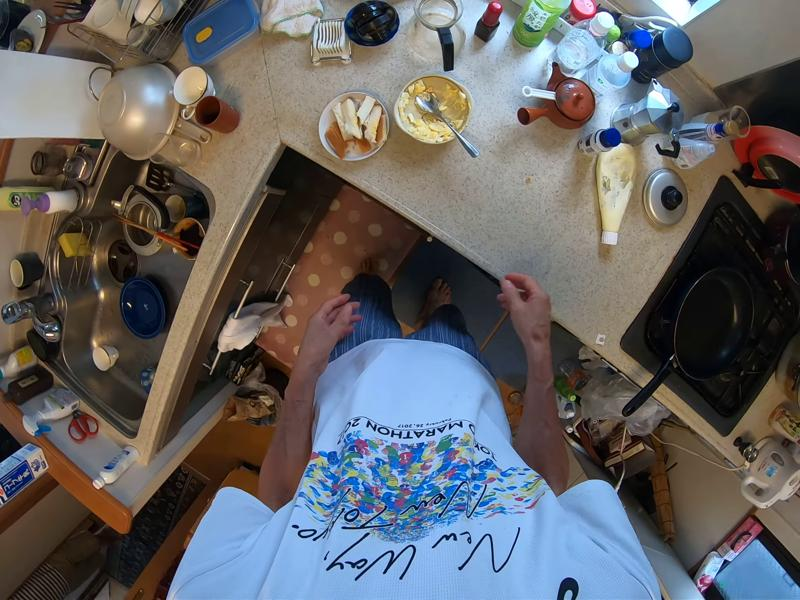} 

        {
        \centering
        
        \textbf{Label}: \textcolor{darkgreen}{\cmark} \hspace{20pt} 
        \textbf{Predicted}: \textcolor{darkred}{\xmark}
        
        }

        \vspace{2pt}

        \textbf{Rationale:} \textit{Is the bottle of mustard...}

        
        \begin{enumerate}[noitemsep,leftmargin=8pt,topsep=0pt,labelsep=2pt]
            \item \textit{on the countertop?} \texttt{Yes}
            \item \textit{on the floor?} \texttt{Yes}
        \end{enumerate}
        
        \end{examplebox}

        \begin{examplebox}{H}{Put the bottle in the cabinet.}{pyellow}

        \includegraphics[width=1\linewidth,trim={80px 45px 40px 90px},clip]{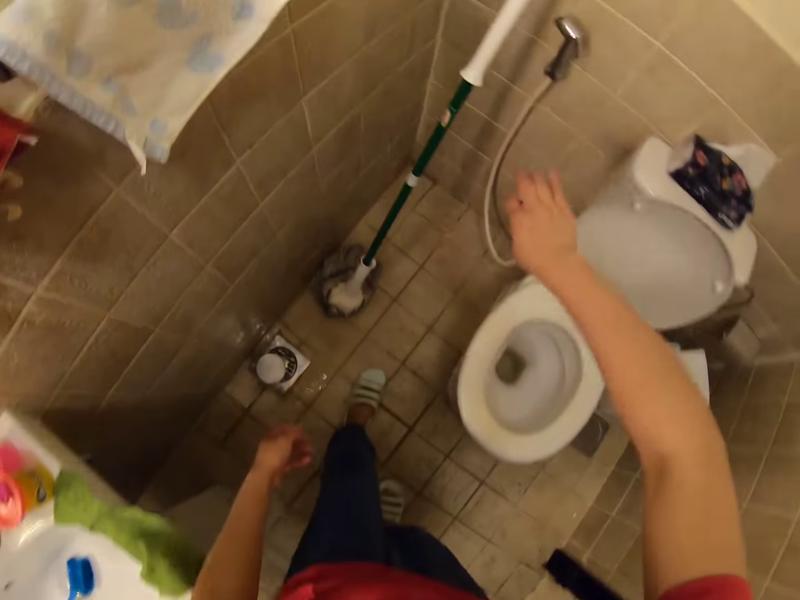} 

        {
        \centering
        
        \textbf{Label}: \textcolor{darkred}{\xmark} \hspace{20pt} 
        \textbf{Predicted}: \textcolor{darkgreen}{\cmark}
        
        }

        \vspace{2pt}

        \textbf{Rationale:} \textit{Is the bottle...}

        
        \begin{enumerate}[noitemsep,leftmargin=8pt,topsep=0pt,labelsep=2pt]
            \item \textit{in the cabinet?} \texttt{Yes}
        \end{enumerate}
        
        \end{examplebox}

    \end{multicols}
    \end{minipage}}

        
    \caption{Sample coherent PMD outputs from LLaVA with coherence-based ranking, representing the range of behaviors observed (as visualized in Figure~\ref{fig:cubes}). Images cropped for visibility and space.}

    
    \label{fig:cube graph examples}
\end{figure*}

\section{Related Work}

Beyond the prior work in PMD discussed in Section~\ref{sec:intro}, we next discuss other related research.

\subsection{Multi-Step Reasoning in LMs \& VLMs}

LMs have exhibited impressive reasoning capabilities from prompting methods \cite{wei-chain-of-thought-2022, kojima2022large}, later strengthened with multiple paths \cite{wang2023selfconsistency, snell2024scaling}, tree-search \cite{yao2023tree, hao-etal-2023-reasoning, putta2024agent, tian2024toward, chen2024alphamath, zhang2024accessing, qi2024mutual}, and fine-tuning on reasoning chains from stronger LMs \cite{wang2023mathcoder,gou2024tora,muennighoff2025s1simpletesttimescaling} and/or self-generated during reinforcement learning \cite{openai2024openaio1card,deepseekai2025deepseekr1incentivizingreasoningcapability}.
Related to PMD, some work has investigated LMs' reasoning about dependencies between physical procedures \cite{bellos-etal-2024-large,lal-etal-2024-cat}. Meanwhile, other work has applied iterative self-questioning approaches to deepen inquiry in other domains, e.g., medicine and fact verification \cite{cattan2024localizingfactualinconsistenciesattributable,li2025aligningllmsaskgood,vladika2025stepbystepfactverificationmedical}.


In light of challenges in visual reasoning in VLMs \cite{dai-etal-2023-plausible,li-etal-2023-evaluating,guan2024hallusionbench}, some work has proposed training-free strategies and training paradigms to reduce visual hallucination \cite{Wan2024CRG,Leng_2024_CVPR_VCD,an2024agla,Ganz_2024_CVPR}, and utilized LMs and other tools to generate intermediate questions and coordinate visual reasoning step by step \cite{you-etal-2023-idealgpt,srinivasan2024selective,chen2024measuring,zhou2023vicor,zong_explainable_audiovisual_pcr,cheng-etal-2024-least,zhu2024chatgpt,sqllava2024,jing2025elevatingvisualquestionanswering,cheng2025visually}.
The latter follows prior work in visual dialog~\cite{Das_2017_CVPR,kottur-etal-2019-clevr}, which has been used to study and improve machine performance in various problem areas \cite{visdial_eval,guesswhat_game,uehara2022,wang-etal-2022-co}. Other works have assessed (V)LMs' reasoning about procedures based on visual information \cite{yang-etal-2018-commonsense,Hendricks2021ProbingIT,jin-etal-2022-leveraging,yuksekgonul2023when,nguyen2024oscar,Nguyen2025NeuroSK}.
In this work, we evaluated self-generated visual dialogs from VLMs in a novel problem of coherent PMD.

\subsection{Leveraging NLI in Other NLP Tasks}

NLI requires judging whether a premise text entails a hypothesis text, a reasoning challenge long studied in NLP \cite{daganPASCALRecognisingTextual2005}. Many human-annotated NLI resources have been created, thus significant progress has occurred in NLI \cite{storks2019recent}. Consequently, prior work has used NLI models to improve the competence, confidence, and coherence of LMs in other tasks, e.g., dialog systems \cite{dziri-etal-2019-evaluating,welleck-etal-2019-dialogue}, summarization \cite{roit-etal-2023-factually}, VQA \cite{srinivasan2024selective}, and image captioning \cite{cascantebonilla2024naturallanguageinferenceimproves}.
We similarly adopt two NLI-based metrics to strengthen VLMs in a new problem of coherent PMD.


%% file: latex/sections/6-conclusion.tex
\section{Conclusion}
In this work, we evaluated foundational VLMs on a challenging problem of coherent PMD, where visual questions and answers must be generated to drive success decisions.
To evaluate these rationales, we leveraged an NLI model to define two coherence metrics, using them to encourage coherent question selection
and generation through common interventions.
Our results showed that VLMs do not generate coherent rationales off-the-shelf, but these interventions improve their coherence, with the former also improving accuracy, and the latter improving the efficiency of generating and extracting information from rationales, albeit creating tradeoffs between these aspects.
Further, patterns in accuracy and coherence metrics revealed detailed performance insights, e.g., visual processing errors like object hallucination.
Ultimately, when choosing an approach for coherent PMD, accuracy, coherence, and efficiency must be weighed for the setting. For example, while online applications may require shorter dialog (even at a cost of accuracy), high-risk applications may prioritize accuracy and confidence. This work lays a foundation for future research in the application of VLMs to PMD and task guidance, and investigation of how these system goals interact. Our code and data are available at \url{https://github.com/sled-group/Transparent-Coherent-PMD}.



%% file: latex/sections/7-etc.tex
\section*{Limitations}\label{sec:limitations}


\paragraph{Latency of self-dialog.}
One limitation of our coherent PMD problem formulation is the requirement of generating several pieces of information autoregressively, which would take several seconds in practical settings (as shown in Appendix~\ref{apx:timing analysis}). This is not ideal for a problem like interactive task guidance, where responsiveness and the ability to intervene quickly to correct mistakes are important.
However, rather than being applied frame by frame (which would likely not be feasible), we expect this process to be applied once at the end of procedure execution to verify the state of the environment, e.g., when the user asks a task guidance system to inform them of the next step of a recipe. Based on the results of our study, one could explore streamlined and specialized approaches to apply VLMs to a stream of video frames in a live online setting. For example, in preliminary experiments, we tried to generate questions once with a VLM, then answer them over a series of video frames, but we found this approach limited by the inability to adapt questions to previously gathered information, and the challenge of aggregating noisy VLM responses across time. We leave further investigation of such approaches for future work.


\paragraph{Inherent limitations of single video frames.}
Next, single video frames are limited in representing actions, which involve movement and state change. Our decision to focus on individual frames stemmed from preliminary experiments we performed with existing open-source VLMs for video understanding \cite{lin2023videollava,li2023otter}, which are still in early stages. There, we found that they often confused information from frames in different segments of the video, preventing them from judging the final states of objects and reconciling this information with success of procedures, thus resulting in poor performance.
As such, our choice to focus on single images simplified the problem for current VLMs, enabling our experiments to begin building a meaningful understanding of their capabilities in PMD. To minimize the dependence on multiple frames in detecting mistakes, we applied several careful preprocessing steps to Ego4D-PMD (as discussed thoroughly in Section~\ref{sec:dataset} and Appendix~\ref{apx: dataset details}). 
It is also worth noting that by not incorporating modalities beyond text and images, e.g., audio, the VLMs we studied are inherently limited in their capturing of physical information \cite{yu2022pacs,zong_explainable_audiovisual_pcr}.

As the ability to reason over sequences of frames and other modalities evolves in state-of-the-art VLMs, future work can revisit this formulation and explore new approaches for them to reason over dynamic scenes. Specifically, we imagine that this task will look somewhat similar as VLMs mature. Given a VLM that can reliably extract information from multiple frames and from audio (e.g., about fine-grained motion, temperatures, and more) and express it in text, coherence metrics like those we explored in this work can still apply. In order to track this information reliably, a neuro-symbolic architecture may be required, e.g., a dynamic scene graph continuously updated across frames based on observations from a VLM and/or other neural vision models. Of course, there will always be some information that a VLM cannot extract from the environment even through vision and audio. For example, confirming amounts or weights of ingredients in a cooking setting may be theoretically possible by reading measurement tools as the human uses them, but this would be a highly difficult challenge. In such cases, a useful capability for a PMD system would be to ask the human user some questions (e.g., “Are you sure that was 3 tablespoons? That looked like a teaspoon.”), then incorporate the user’s answers into PMD and rationalization. Such a capability becomes a rich research problem of its own, as it would be important for the system not to ask too many questions \cite{bao-etal-2023-foundation}, requiring it to be able to identify and prioritize the cases where this is necessary, and to ask questions concisely and in a timely manner.

\paragraph{Lack of ground truth rationales.}
Lastly, it is worth noting that the dataset collected does not include ground truth rationales for mistake detection labels.
Instead, we opted to propose automated coherence metrics for generated questions and answers based on a fine-tuned NLI model, which itself is prone to error and thus limits the objectivity of our evaluations. We chose this path for two important reasons. First, there may be multiple valid ways to detect a mistake through asking and answering visual questions, each of which could consist of different questions and/or different numbers of questions.\footnote{For example, in trying to determine the success of the procedure ``In a bowl, add the cut cherry tomatoes'' from \citet{peddi2023captaincook4d}, we could reasonably ask one question ``Are all the cherry tomatoes in the bowl?'' or two questions ``Are there cherry tomatoes in the bowl?'' and ``Are there any cherry tomatoes outside of the bowl?''} In our opinion, existing metrics for text generation (which largely measure syntactic or semantic similarity of text) are not as well suited for this extremely challenging evaluation as fine-tuned NLI models (which are optimized to judge the logical consistency between texts). Second, in a real-world setting like PMD for task guidance, we believe that automated metrics are better suited for continually understanding and improving a deployed system than an offline benchmark of ground truth rationales. Despite this limitation, we believe that the value of these metrics is demonstrated by the accuracy and efficiency improvements brought by incorporating them into coherent PMD.

\section*{Acknowledgments}
This work was supported in part by the DARPA PTG program HR00112220003, NSF IIS-1949634, ARPA-H PARADIGM program 1AY2AX000062, the Microsoft Accelerate Foundation Models Research (AFMR) grant program, and computational resources and services provided by Advanced Research Computing at the University of Michigan, Ann Arbor. We would like to thank our anonymous reviewers, as well as Megan Su, Ruixuan Deng, Fengyuan Hu, Andy Chung, Lu Wang, Richard L. Lewis, and the entire MSRP team, for their helpful discussions, feedback, and insights. ChatGPT\footnote{\url{https://chatgpt.com/}} was used for minor writing suggestions and tedious coding tasks (e.g., improving generated graphs). This research was funded, in part, by the U.S. Government. The views and conclusions contained in this document are those of the authors and should not be interpreted as representing the official policies, either expressed or implied, of the U.S. Government.

%% file: latex/sections/8-apx.tex
\section{Ego4D-PMD Data Curation Details}\label{apx: dataset details}

Various benchmark datasets have been created for PMD from egocentric video, each of which includes video and procedural text along with various other modalities, as well as detailed information about mistakes \cite{pmlr-v232-du23b,bao-etal-2023-foundation,HoloAssist2023,peddi2023captaincook4d}.
While these datasets are useful resources for research in task guidance, they are video-based and cover limited domains. Further, most of them include dialog interaction between a user and instructor agent which often causes mistakes to be corrected before or while they happen. They include mistakes around temperature, timing, small measurements, and other physical properties of the environment that are difficult for open-source VLMs, which are mostly optimized for representing single images, to perceive. While some of these issues could be overcome through a two-way dialog between the agent and a user, this makes it harder to isolate mistakes occurring in the videos and dive deep into the reasoning behind detecting them. 

To alleviate these challenges and focus our inquiry, we follow \citet{pmlr-v232-du23b} in recasting Ego4D, a procedural video dataset with a breadth of annotated and narrated everyday actions, into a single frame mistake detection format.
The collection of the Ego4D~\cite{Ego4D2022CVPR} for Procedural Mistake Detection (Ego4D-PMD) data consisted of several preprocessing steps, outlined below.

\paragraph{Generating success examples.}
As discussed above, Ego4D's hand object interaction data is annotated in units of egocentric video clips of individual actions being performed by humans. We can form an example of a successful execution of the procedure by pairing each video clip with its annotated natural language narration of the procedure. Since most VLMs are not optimized to reason over multiple frames and videos, and those that are are still in very early stages, we sample exactly one frame from each video clip. Specifically, as each clip is carefully annotated with a postcondition time for the action, i.e., the time that the action has been completed, we simply sample the video frame at this annotated time and pair it with the text narration.

\paragraph{Generating mistake examples for incomplete procedures.}
One natural type of mistake a user could make is not finishing a procedure. In addition to postcondition times, each video clip is annotated with a precondition time. Following a similar approach to \citet{pmlr-v232-du23b}, we can generate a mistake example by sampling a frame at the precondition time and pairing it with the video clip's narration text. We expect that by doing this, the sampled frame will show the procedure at an incomplete state, and contain most of the same objects as the success example for the same clip. This poses a difficult challenge of identifying the key physical properties of the scene that would indicate completion.

\paragraph{Generating mistake examples for mismatched verbs and nouns.}
Mistakes also happen when a user applies the wrong type of action to an object, causing an unexpected state, as well as when a user uses the wrong object or ingredient in a procedure. Following this intuition, we generate additional easier mistake examples from each clip by matching each clip with other clips that have a mismatched verb, noun, or both. While each clip is annotated with verb and noun categories, these categories are coarse-grained, making it impossible to guarantee that two clips with the same verb or noun label actually involve the same verb or noun, thus preventing sampling clips that share the same verb or noun. Instead, we apply the \texttt{AllenNLP}\footnote{\url{https://allenai.org/allennlp}} semantic role labeler to each narration text to identify the key participants in each procedure. For each clip, we then attempt to sample the postcondition frames from three mismatched clips: one with a mismatched verb (but matching nouns), one with a mismatched noun (but matching verb), and one with a mismatched verb and noun. To avoid incorrect mismatches, we omitted some verbs from this matching that are often prerequisite for other object interactions and thus likely to occur even in sampled videos thought to have mismatched verbs, such as movement verbs like “take” and “put.” We also verified that the verb-noun pair from the source video did not happen earlier in the retrieved video, ensuring that the correct physical states cannot be observed in the retrieved video from earlier actions.
We then pair sampled frames with the source clip's narration text, creating mistake examples with varying levels of overlap with the source clip. While it is not always possible to find every such alternative clip for each clip in Ego4D, we can usually find at least one of them. 

\paragraph{Transforming narrations into instructions.}
The narration texts annotated in Ego4D are declarative statements about the actions being performed in each clip. This is not an accurate depiction of typical interactive task guidance and PMD settings, which usually revolve around instructional texts like recipes or guidebooks. As such, we convert each narration, e.g., ``Someone washes the lettuce,'' into imperative form, e.g., ``Wash the lettuce,'' using \texttt{spaCy}.\footnote{\url{https://spacy.io/}} Further, some narrations describe procedures that are not suited for comparing physical state changes in text and images, such as social interactions, interactions with animals, interactions with electronic devices, and movements that are impossible to precisely characterize from the narration text (e.g., in ``Move plate''). We use the verb and noun category annotations on each clip to filter out such cases.

\paragraph{Ensuring data quality.}
We perform several additional steps to ensure high-quality mistake detection task instances. First, we remove clips where the precondition and postcondition frames are overly similar (i.e., at least 0.95 cosine similarity). We remove clips that are too dark (i.e., where the mean of all normalized RGB values is less than 0.2). When sampling frames from source clips, we sample several candidates within a small range around the precondition or postcondition timestamp, then select the least blurry candidate by the variance of the images' Laplacian. Some videos in Ego4D show the same action being performed repeatedly (e.g., ``Roll a ball of dough''), which can make it difficult to determine whether the state of the environment shown in a clip is the result of the current procedure or a prior one (given only a single frame). While future work applying video-optimized VLMs for coherent PMD in long-horizon tasks will need to address this challenge, this adds an unnecessary complexity to an already challenging task for current VLMs. As such, we remove any clips such that the same procedure in the clip has already been performed previously in the video. Lastly, we remove a small number of videos in Ego4D that we notice to be corrupted or significantly distorted.

Various statistics of the resulting full Ego4D-PMD dataset (and sub-samples used in the paper) are presented in Tables~\ref{tab:ego4d md stats} and \ref{tab:ego4d md stats 2}. 


\begin{table}
    \centering
    \footnotesize
    \setlength{\tabcolsep}{3pt}
    \scalebox{0.76}{
    
    \begin{tabular}{cccccccc}\toprule
        \textbf{Type} & \textbf{Train} & (Sample) & \textbf{Val.} & (Sample) & \textbf{Test} & (Sample) & \textbf{Total}  \\\cmidrule(lr){1-1}\cmidrule(lr){2-3}\cmidrule(lr){4-5}\cmidrule(lr){6-7}\cmidrule(lr){8-8}
        {Success} & 42.0k & 5.00k & 13.1k & 250 & 18.1k  & 1.00k & 73.1k \\\cmidrule(lr){1-1}\cmidrule(lr){2-3}\cmidrule(lr){4-5}\cmidrule(lr){6-7}\cmidrule(lr){8-8}
        Mistake & 99.4k & 5.00k & 25.4k & 250 & 34,182 & 1.00k & 159k \\\cmidrule(lr){1-1}\cmidrule(lr){2-3}\cmidrule(lr){4-5}\cmidrule(lr){6-7}\cmidrule(lr){8-8}
        \textit{(Incomplete)} & 15.1k & 755 & 4.91k & 51 & 6.55k  & 194 &  26.5k \\
        {\textit{(Wrong V)}} & 11.8k & 604 & 2.69k & 31 & 3.75k  & 108 & 18.2k \\
        {\textit{(Wrong N)}} & 36.4k & 1.85k & 8.91k & 87 & 11.8k  & 344 & 57.2k \\
        {\textit{(Wrong V\&N)}} & 36.1k & 1.79k & 8.91k & 81 & 12,047 & 354 & 57.1k \\\bottomrule
    \end{tabular}
    
    }
    
    \caption{Distribution of example types in each partition of our curated Ego4D \cite{Ego4D2022CVPR} for Procedural Mistake Detection (Ego4D-PMD) dataset.}
    \label{tab:ego4d md stats}
\end{table}

\begin{table}
    \centering
    \footnotesize
    \setlength{\tabcolsep}{2pt}
    \scalebox{0.87}{
    
    \begin{tabular}{cccccccc}\toprule
        \textbf{Type} & \textbf{Train} & (Sample) & \textbf{Val.} & (Sample) & \textbf{Test} & (Sample) & \textbf{All} \\\cmidrule(lr){1-1}\cmidrule(lr){2-3}\cmidrule(lr){4-5}\cmidrule(lr){6-7}\cmidrule(lr){8-8}
        Verbs & 83 & 80 & 77 & 55 & 78 & 71 & 83 \\\cmidrule(lr){1-1}\cmidrule(lr){2-3}\cmidrule(lr){4-5}\cmidrule(lr){6-7}\cmidrule(lr){8-8}
        Nouns & 440 & 372 & 365 & 151 & 390 & 257 & 487 \\\cmidrule(lr){1-1}\cmidrule(lr){2-3}\cmidrule(lr){4-5}\cmidrule(lr){6-7}\cmidrule(lr){8-8}
        V-N Pairs & 3,976 & 2,050 & 2,185 & 326 & 2,658 & 833 & 5,363 \\\bottomrule
    \end{tabular}
    
    }
    
    \caption{Distribution of unique verb, noun, and verb-noun pair categories in each partition of our curated Ego4D-PMD dataset. Verb and noun categories are annotated for each narration in the Ego4D dataset.}
    \label{tab:ego4d md stats 2}
\end{table}

\section{Rephrasing Questions and Answers for Coherence Evaluation}\label{apx: pmd rephrase}
As discussed in Section~\ref{sec: pmd nli}, we use a fine-tuned NLI model to judge the success of procedures given questions and answers. In order to convert questions and answers into declarative statements to pass into the NLI model, we follow \citet{srinivasan2024selective} in prompting a VLM with the following 10 in-context demonstrations of rephrasing before prompting it to rephrase a question and answer for the task at hand:

\begin{enumerate}[label=\arabic*.]
    \item \textit{\textbf{Question:}} \textit{Is there a bowl on the table?} \\
    \textit{\textbf{Answer:}} \textit{Yes} \\
    \textit{\textbf{Statement:}} \textit{There is a bowl on the table.}
    
    \item \textit{\textbf{Question:}} \textit{Are the eggs cracked?} \\
    \textit{\textbf{Answer:}} \textit{No} \\
    \textit{\textbf{Statement:}} \textit{The eggs are not cracked.}
    
    \item \textit{\textbf{Question:}} \textit{Does the cardboard box look open?} \\
    \textit{\textbf{Answer:}} \textit{Yes} \\
    \textit{\textbf{Statement:}} \textit{The cardboard box looks open.}
    
    \item \textit{\textbf{Question:}} \textit{Are there any leaves outside of the basket?} \\
    \textit{\textbf{Answer:}} \textit{No} \\
    \textit{\textbf{Statement:}} \textit{There are not any leaves outside of the basket.}
    
    \item \textit{\textbf{Question:}} \textit{Is the orange peeled?} \\
    \textit{\textbf{Answer:}} \textit{Yes} \\
    \textit{\textbf{Statement:}} \textit{The orange is peeled.}
    
    \item \textit{\textbf{Question:}} \textit{Is the mug empty?} \\
    \textit{\textbf{Answer:}} \textit{No} \\
    \textit{\textbf{Statement:}} \textit{The mug is not empty.}
    
    \item \textit{\textbf{Question:}} \textit{Are there hedge trimmers in the image?} \\
    \textit{\textbf{Answer:}} \textit{Yes} \\
    \textit{\textbf{Statement:}} \textit{There are hedge trimmers in the image.}
    
    \item \textit{\textbf{Question:}} \textit{Has the light switch been turned on?} \\
    \textit{\textbf{Answer:}} \textit{No} \\
    \textit{\textbf{Statement:}} \textit{The light switch has not been turned on.}
    
    \item \textit{\textbf{Question:}} \textit{Does the table have any cups on it?} \\
    \textit{\textbf{Answer:}} \textit{Yes} \\
    \textit{\textbf{Statement:}} \textit{The table has cups on it.}
    
    \item \textit{\textbf{Question:}} \textit{Is the cabinet closed?} \\
    \textit{\textbf{Answer:}} \textit{No} \\
    \textit{\textbf{Statement:}} \textit{The cabinet is not closed.}
\end{enumerate}

\section{Human Annotation Details}

In this appendix, we collect human judgments for the relevance and informativeness metrics defined in Section~\ref{sec:pmd metrics}, as well as PMD decisions. 

\subsection{Validating Coherence Metrics}\label{apx: pmd human study details}

To achieve this, we randomly sampled the outputs for 50 iterations of LLaVA's self-dialog from two combinations of approaches:
\begin{itemize}
    \item Likelihood-based question ranking
    \item Coherence-based question ranking augmented with question candidates from in-context learning
\end{itemize}

These outputs were from intermediate experiments, and thus the prompt used to initiate the self-dialog is slightly different than the one used for the experiments in the paper. However, as this is just a source of questions to compare human and machine judgments, this discrepancy does not impact the conclusions in the paper.

\subsubsection{Annotation Instructions}
For both relevance and informativeness annotation, we provided the following background for the task: 

\begin{quote}
    \textit{Imagine you just had eye surgery, and are currently unable to see. You're performing a task you're familiar with, but need help to determine whether you successfully completed it. You video call a friend (who is unfamiliar with the task) and show them what you're working on. You then ask them some yes/no questions to figure out whether you successfully completed the task.}    
\end{quote}

\paragraph{Relevance annotation instructions and example.}

Annotators are provided the following instructions for annotating relevance:

\begin{quote}
    \textit{For each annotation task, you will be given the following information:}
    \begin{itemize}[topsep=0pt]
        \item \textit{A \textbf{sentence} describing the procedure you're trying to perform.}
        \item \textit{An optional list of \textbf{previous questions} you already asked, and their \textbf{answers}.}
        \item \textit{A \textbf{potential next question} you could ask your friend.}
    \end{itemize}

    \textit{You must rate how \textbf{relevant} the potential next question is. By relevant, we mean: \textbf{given the previous questions and answers, how helpful could an answer to this question be in determining whether you successfully completed the task?}}

    \textit{You can also choose to mark "Instructions Unclear", which means that the sentence itself is not clear, so you're not sure how to determine whether the procedure is successful. This should only be used in rare cases.}

    \textbf{\textit{Some tips:}}
    \begin{itemize}
        \item \textit{Only judge the relevance of the potential next question, not the previous questions (which may or may not be relevant).}
        \item \textit{A question may seem relevant to the task at hand, but you should consider it irrelevant if it can't provide essential information to judge whether the task was completed successfully.}
        \item \textit{If a seemingly relevant question is redundant with previous questions, you may consider it less relevant.}
        \item \textit{Assume that the answer to the question won't contradict the information you have from previous questions and answers. If previous questions and answers already contradict each other, consider whether this question could sway you one way or another.}
        \item \textit{The instructional text and questions may refer to "someone" or "a person"; always assume this is referring to yourself (the person performing the task).}
        \item \textit{The questions may refer to a "photo" or "image"; always assume this is referring to the video feed your friend would see through the video call.}
    \end{itemize}

\end{quote}

One sample is listed below:

\begin{quote}
    \textit{\textbf{Sentence:} Drop the bowls on the table with your hand}

    \textbf{\textit{Previous questions and answers:}}

    \begin{enumerate}[topsep=0pt]
        \item \textit{Are the bowls on the table? (Answer: Yes)}
        \item \textit{Is the person holding the bowls in their hand? (Answer: No)}
    \end{enumerate}

    \textit{\textbf{Potential next question:}} \textit{Is the person about to drop the bowls on the table?}

    \textit{\textbf{Your rating:}}

    \begin{itemize}[topsep=0pt]
        \item \textit{1 (very irrelevant)}
        \item \textit{2 (slightly irrelevant)}
        \item \textit{3 (neutral; may or may not be relevant)}
        \item \textit{4 (slightly relevant)}
        \item \textit{5 (very relevant)}
        \item \textit{Instructions Unclear}
    \end{itemize}
\end{quote}

\paragraph{Informativeness annotation instructions and example.}

Annotators are provided the following instructions for annotating informativeness:

\begin{quote}
    \textit{For each annotation task, you will be given the following information:}
    \begin{itemize}[topsep=0pt]
        \item \textit{A \textbf{sentence} describing the procedure you're trying to perform.}
        \item \textit{A list of \textbf{questions} you asked your friend, and their \textbf{answers}.}
    \end{itemize}

    \textit{You must rate how \textbf{informative} the questions and answers are. By informative, we mean: \textbf{based on all the information you have, how sure are you about whether you succeeded?}}

    \textit{You can also choose to mark "Instructions Unclear", which means that the sentence itself is not clear, so you're not sure how to determine whether the procedure is successful. This should only be used in rare cases.}

    \textit{\textbf{Some tips:}}
    \begin{itemize}
        \item \textit{Your task is to rate how sure you are, NOT whether you believe the procedure is successfully completed or not.}
        \item \textit{Consider all questions and answers as a whole; if you have contradictory information, this may reduce your sureness.}
        \item \textit{The instructional text and questions may refer to "someone" or a "person"; always assume this is referring to yourself (the person performing the task).}
        \item \textit{The questions may refer to a "photo" or "image"; always assume this is referring to the video feed your friend would see through the video call.}
    \end{itemize}

\end{quote}

One sample is listed below:

\begin{quote}
    \textit{\textbf{Sentence:} Clean the bowl}

    \textbf{\textit{Previous questions and answers:}} \textit{None}

    \textit{\textbf{Last question:}} \textit{Is there a bowl in the image?}
    \textit{\textbf{Last answer:}} \textit{Yes}

    \textit{\textbf{Your rating:}}

    \begin{itemize}[topsep=0pt]
        \item \textit{1 (very uninformative/unsure)}
        \item \textit{2 (slightly uninformative/unsure)}
        \item \textit{3 (neutral; may or may not be relevant)}
        \item \textit{4 (slightly informative)}
        \item \textit{5 (very informative)}
        \item \textit{Instructions Unclear}
    \end{itemize}
\end{quote}

\subsubsection{Annotation Results}
For each metric, we recruited 5 annotators (all English speakers with conferred or in-progress undergraduate degrees) to rate the relevance and informativeness of 10 machine-generated questions and answers. As this was an initial pilot study that was not scaled up further, annotators were authors of the work (not the first author) and/or peers of the authors, and not compensated. When comparing with our automated relevance and informativeness metrics, we use LLaVA for rephrasing.

\paragraph{Comparison of automated and human relevance judgments.}
We presented each annotator with 10 randomly selected VLM-generated questions $Q'$, along with previous questions and answers $\mathcal{Q}$ and $\mathcal{A}$. Annotators were instructed to rate the relevance (i.e., given the previous questions and answers, how helpful could an answer to this question be in determining whether the task was successfully completed) on a scale from 1-5 (least to most relevant). Between the resulting 50 annotations and corresponding automated metrics, we found a moderate Spearman correlation \cite{spearman1961proof} of $\rho = 0.55$ ($p=0.000037$). This suggests that this automated measure of relevance is indeed positively correlated with human judgments of relevance.

\paragraph{Comparison of automated and human informativeness judgments.}
We presented each annotator with 10 randomly selected VLM-generated questions $Q'$ and answers $A'$, along with previous questions and answers $\mathcal{Q}$ and $\mathcal{A}$. Annotators were instructed to rate the relevance (i.e., based on all the information we have, how sure is the annotator about whether the procedure was successfully completed) on a scale from 1-5 (least to most informative). Between the resulting 50 annotations and corresponding automated metrics, we found a weaker Spearman correlation \cite{spearman1961proof} of $\rho = 0.33$ ($p=0.020$). Interestingly, if we multiply the automated informativeness metrics by the relevance for $Q'$, $\mathcal{Q}$, and $\mathcal{A}$, we find a stronger Spearman correlation of $\rho = 0.50$ ($p=0.00022$). This suggests that while informativeness does have a relationship with human judgments, when multiplying it by relevance this relationship is stronger and more significant. This might be because the concepts of relevance and informativeness are themselves related. Intuitively, in most cases, a relevant question should be informative, and an irrelevant question should be uninformative. Meanwhile, fine-tuned NLI models could theoretically score an answer to an irrelevant question as informative, and vice versa. When proposing a coherence-based question re-ranking strategy in Section~\ref{sec:coherence ranking}, we incorporated the inductive bias reflected in these human judgments by multiplying informativeness by relevance. Later, in Section~\ref{sec: final analysis}, we visualized the full distribution of relevance and informativeness on our evaluation data to better understand this issue.

\subsection{Human PMD Accuracy}\label{apx:human accuracy annotation}

To create a reference point for PMD accuracy measurements of VLMs and a proxy for the Ego4D-PMD data quality and objectiveness, we performed a small human annotation study for the task of PMD. Specifically, we randomly selected 100 examples from the testing data, and asked 3 annotators to judge whether each example's frame showed a successful execution of the actions described by its procedural text. We recruited a total of 12 annotators, each of which labeled 25 examples. All annotators were proficient English speakers with conferred or in-progress undergraduate degrees, and again were uncompensated authors of the work (not the first author) and/or peers of the authors. Annotators were given the following instructions:

\begin{quote}
    \textbf{You will be shown 25 pairs of task instructions and egocentric (POV) photos from people's perspective. Based on each photo, you will be asked whether the person has successfully completed the given task.} You might see a range of situations, including the person following the instructions perfectly, making a minor mistake, or doing something completely different.

\textbf{Some guidelines:}

\begin{itemize}
    \item Based on the task instructions, you should look for one or more objects that you expect to be involved in the task. Based on what they look like and where they're located, you should judge whether the task has been successfully completed.
    
    \begin{itemize}
        \item For example, if the task is to "slice an apple," and you only see whole unsliced apples in the photo, it should be labeled a failure ("No - They did something wrong").
        \item Or if you do see a sliced apple, it should be labeled a success ("Yes - Task completed successfully").
    \end{itemize}

    \item It's possible the photo doesn't provide enough information to decide if the task is complete. If you don't see any objects required for the task, only see part of an object, or the photo is blurry/low quality, it may or may not still be a success. Make your best guess based on what you do see in the image.

    \begin{itemize}
        \item You will also have a chance to indicate your confidence.
    \end{itemize}

\end{itemize}

\end{quote}

For each example, annotators were shown the example's video frame and procedural text, then asked:

\begin{quote}
    \textbf{Did the person successfully complete the task?}
    \begin{itemize}
        \item No - They did something wrong
        \item Yes - Task completed successfully 
    \end{itemize}
\end{quote}

They were also asked to indicate their confidence on a scale from 1 to 5:

\begin{quote}
    \textbf{How confident are you in your answer?}
    \begin{itemize}
        \item 1 - Not confident at all
        \item 2 - Slightly confident
        \item 3 - Moderately confident
        \item 4 - Very confident
        \item 5 - Extremely confident
    \end{itemize}
\end{quote}

Pairs of questions for examples were presented in a random order for each annotator. To calculate human accuracy, we explored two approaches: (1) calculating the accuracy over the 3 annotators for each example then averaging over all examples, and (2) taking the majority judgement from the 3 annotators then calculating the accuracy based on those. The mean human accuracy was 71.3\%, and the majority human accuracy was 72.0\%, compared to the best model result of 67.8\% in the coherent PMD setting.\footnote{GPT-4o achieves 69.2\% accuracy in the rationale-free evaluation presented in Appendix~\ref{apx:exp free results}, which is closer to human performance. However, it is important to note that this evaluation excludes the step of forming a rationale, which we assume humans do implicitly in making their decisions.}
This shows that VLMs still lag slightly behind humans in accurately detecting mistakes, and if human performance is to be considered an upper bound, coherently rationalizing decisions will remain a crucial step in this problem. Meanwhile, the mean human confidence rating was 3.52 out of 5, suggesting humans were usually moderately to very confident in their decisions. Lastly, the mean Cohen's $\kappa$ \cite{cohen} for inter-annotator agreement was 0.572, suggesting moderate agreement on decisions.

\section{VLM Self-Dialog Details}\label{apx: pmd prompt details}

In this appendix, we include prompt templates and other supplementary details for the self-dialog we conditioned VLMs to generate, including visual question generation, visual question answering, and stopping criteria. To conserve GPU memory, VLM weights are 4-bit quantized at inference time, and are obtained from \url{https://huggingface.co/Salesforce/instructblip-vicuna-7b}, \url{https://huggingface.co/liuhaotian/llava-v1.5-7b}, and \url{https://huggingface.co/meta-llama/Llama-3.2-11B-Vision}.

\subsection{Visual Question Generation}\label{apx: vqg prompt details}
When prompting vision-and-language models (VLMs) to generate questions, we use the following prompt template:

\begin{quote}

    \textit{This is a photo of someone working on the procedure ``$\langle$procedural text$\rangle$''. I will ask a series of different yes/no questions to gather information about the state of the scene, then use it to determine whether the person has successfully completed the procedure. The goal is to extract as much relevant information as possible from the scene, so I will not repeat questions. I will try to ask short and simple questions about physical states and locations that are possible to observe from the photo.}
    
    \textit{Q:}
\end{quote}

Question generation is not conditioned on video frames, as we found significant performance degradation when VLMs were conditioned on the video frame while generating questions, often leading to completely nonsensical questions, e.g., ``Is is is is is is?'' To ensure the VLM generates yes-no questions, we constrain generation during decoding to enforce that each generated text begins with a word that can signal a yes-no question,\footnote{Specifically, questions must begin with \textit{is}, \textit{does}, or \textit{has}, along with all plural and past tense forms of these verbs.}
does not include the word \textit{or}, and ends with a question mark. To avoid vague and high-level questions about the status of the procedure rather than low-level physical states, we prevent VLMs from repeating some words present in their input prompts: ``successful,'' ``successfully,'' ``completed,'' and ``procedure.'' 
To encourage logical questions while ensuring variety, we apply greedy beam search decoding with $k=8$ beams, returning the top 4 candidate questions to be ranked through LM likelihood or coherence metrics.\footnote{Due to generation constraints, it is often the case that the VLM does not successfully generate all 8 candidates. In rare cases, these constraints prevent VLMs from generating any candidate questions; when this happens, we repeat the generation without any constraints (even though this may result in outputs that are not yes-no questions).}
Out of the candidates, we remove any that are exactly the same as previously generated questions, then select the most likely candidate based on the model's log-likelihoods. 

\subsection{Visual Question Answering}
Once a question is generated, we prompt the VLM with it and the video frame along with ``A:'' to elicit an answer. To produce the answer, we apply softmax over the output logits from the forward pass of the VLM for the \textit{Yes} and \textit{No} tokens. It is important to note that we exclude the dialog history from the context during VQA, as we again observed significant performance degradation when VLMs answered visual questions in the context of a longer dialog. This was especially prominent when several similar questions were generated and answered in a dialog, which often caused the VLM to creep from being initially unsure about the answer to being confidently wrong.

\subsection{Success Classification}
Questions and answers are generated iteratively until the stopping criteria described in Section~\ref{sec:stopping criteria} are met. To prompt the VLM to judge the success of a procedure after a question is answered, we use the following prompt:

\begin{quote}
    \textit{Q: Based on the image and above information, has the procedure ``$\langle$procedural text$\rangle$'' been successfully completed? A:}
\end{quote}

Here, the logits of the \textit{Yes} and \textit{No} tokens are similarly used to produce a probability distribution over a success or mistake decision. The decision is determined by a mistake confidence threshold $\tau$, which is selected to maximize overall accuracy from a comprehensive set of 99 candidates $\tau \in \{0.01, 0.02, \dots, 0.98, 0.99, 1.0 \}$.


\subsection{Stopping Criteria Details}\label{apx:stop criteria details}
The stopping criteria hyperparameter $n^*$ is fixed at a value of 10. We chose to fix $n^*$ because tuning $n^*$ prevents objective comparison of the number of iterations taken by each approach, and $\delta$ and $\epsilon$ can still control the number of iterations of VLMs' self-dialog. The hyperparameters $\delta$ and $\epsilon$ are selected from a grid search on the validation data over combinations of $\delta \in \{ 0.05, 0.1, 0.2, 0.4 \}$ and $\epsilon \in \{ 0.025, 0.05, 0.1, 0.2 \}$. We maximize overall accuracy, relevance of questions, and potential informativeness of their answers. Specifically, we use a cascading summary metric which is the product of example-level informativeness (reference-adjusted) and example-level relevance if the VLM makes a correct mistake detection classification and this product is positive, else it is zero. Accuracy depends on the mistake confidence threshold $\tau$, which is selected as described in the previous subsection.
Selected stopping criteria hyperparameter values for the results in Table~\ref{tab:selection metrics} are listed later in Appendix~\ref{apx:selection hyperparam}, while values for the results in Table~\ref{tab:generation metrics} are listed in Appendix~\ref{apx:dpo details}.

\section{Supplementary Experimental Details and Results}\label{apx:supp results}

In this appendix, we provide assorted details (e.g., prompt templates and hyperparameters) and supplementary results that were omitted from the main body of the paper.

\subsection{Example Questions for In-Context Learning}\label{apx: pmd icl}

As discussed in Section~\ref{sec:pmd icl details}, we condition VLMs with sets of human-written questions for 20 procedures from the Ego4D for Procedural Mistake Detection (Ego4D-PMD) dataset. These human-written questions individually achieve 53.2\% relevance and 82.1\% maximum informativeness (i.e., for a \textit{yes} or \textit{no} answer) on average, with rephrasing done by LLaVA. The annotated procedures (underlined) and questions (italicized) are listed below:

\begin{enumerate}[noitemsep]
    \item \ul{Soak the sponge in a soapy water with your hands}
    \begin{enumerate}[noitemsep]
        \item \textit{Is there a sponge?}
        \item \textit{Is the sponge in water?}
        \item \textit{Is the water soapy?}
    \end{enumerate}
    \item \ul{Open the bottle}
    \begin{enumerate}[noitemsep]
        \item \textit{Is there a bottle in the image?}
        \item \textit{Is the bottle open?}
        \item \textit{Does the bottle have a lid on it?}
    \end{enumerate}
    \item \ul{Take the baking tray away from the table}
    \begin{enumerate}[noitemsep]
        \item \textit{Can you see a baking tray?}
        \item \textit{Is the baking tray on the table?}
        \item \textit{Is the baking tray picked up by someone?}
    \end{enumerate}
    \item \ul{Turn on a torch light}
    \begin{enumerate}[noitemsep]
        \item \textit{Is there a torch light in the photo?}
        \item \textit{Is the torch light powered on?}
        \item \textit{Is the torch light lit up?}
    \end{enumerate}
    \item \ul{Fold the right edge of the wrapper}
    \begin{enumerate}[noitemsep]
        \item \textit{Is there a wrapper in the image?}
        \item \textit{Is the wrapper completely flat?}
        \item \textit{Is the right edge of the wrapper folded?}
    \end{enumerate}
    \item \ul{Pour the water into the blue container}
    \begin{enumerate}[noitemsep]
        \item \textit{Do you see a blue container anywhere?}
        \item \textit{Is there water in the blue container?}
        \item \textit{Is the blue container empty?}
    \end{enumerate}
    \item \ul{Paint the patio with the paint brush}
    \begin{enumerate}[noitemsep]
        \item \textit{Is this a photo of a patio?}
        \item \textit{Is the patio painted?}
        \item \textit{Is someone holding a paint brush?}
    \end{enumerate}
    \item \ul{Spread the black peas on the salad with the spoon in your hand}
    \begin{enumerate}[noitemsep]
        \item \textit{Is there a salad?}
        \item \textit{Are there black peas on the salad?}
        \item \textit{Is there a spoon in someone's hand?}
    \end{enumerate}
    \item \ul{Scoop paint from the pallet on the table with the paint brush}
    \begin{enumerate}[noitemsep]
        \item \textit{Do you see a paint brush and a paint palette?}
        \item \textit{Is there paint on the paint brush?}
        \item \textit{Is the paint brush in someone's hand?}
    \end{enumerate}
    \item \ul{Wash the car with a sponge in your hand}
    \begin{enumerate}[noitemsep]
        \item \textit{Do you see a car?}
        \item \textit{Is the car clean?}
        \item \textit{Is the sponge being held?}
    \end{enumerate}
    \item \ul{Pick the scrubber from the sink}
    \begin{enumerate}[noitemsep]
        \item \textit{Do you see a scrubber somewhere?}
        \item \textit{Is the scrubber in the sink?}
        \item \textit{Is the scrubber in someone's hand?}
    \end{enumerate}
    \item \ul{Peel the onion}
    \begin{enumerate}[noitemsep]
        \item \textit{Is there an onion in the image?}
        \item \textit{Is the onion's skin removed?}
        \item \textit{Is the onion peeled?}
    \end{enumerate}
    \item \ul{Put the dirt in the dust bin}
    \begin{enumerate}[noitemsep]
        \item \textit{Is there a dust bin?}
        \item \textit{Is there dirt in the dust bin?}
        \item \textit{Is there any dirt outside of the dust bin?}
    \end{enumerate}
    \item \ul{Cut dough into two}
    \begin{enumerate}[noitemsep]
        \item \textit{Do you see any dough?}
        \item \textit{Is the dough in two pieces?}
        \item \textit{Is the dough whole?}
    \end{enumerate}
    \item \ul{Break the walnut with the nutcracker in your hand}
    \begin{enumerate}[noitemsep]
        \item \textit{Do you see a walnut?}
        \item \textit{Is the walnut cracked?}
        \item \textit{Is there a nut cracker in someone's hand?}
    \end{enumerate}
    \item \ul{Turn off the tap}
    \begin{enumerate}[noitemsep]
        \item \textit{Is there a tap in the photo?}
        \item \textit{Is the water running?}
        \item \textit{Is the faucet switched off?}
    \end{enumerate}
    \item \ul{Heat the edge of the bag with the lighter}
    \begin{enumerate}[noitemsep]
        \item \textit{Do you see a bag and a lighter?}
        \item \textit{Is there a flame coming from the lighter?}
        \item \textit{Is the lighter near the bag?}
    \end{enumerate}
    \item \ul{Close the fridge}
    \begin{enumerate}[noitemsep]
        \item \textit{Is there a fridge?}
        \item \textit{Is the fridge open?}
        \item \textit{Can you see inside the fridge?}
    \end{enumerate}
    \item \ul{Chop green beans with a knife on the chopping board}
    \begin{enumerate}[noitemsep]
        \item \textit{Do you see green beans on a cutting board?}
        \item \textit{Are the green beans sliced?}
        \item \textit{Is someone using a knife?}
    \end{enumerate}
    \item \ul{Drop the brush in your hand on the oven}
    \begin{enumerate}[noitemsep]
        \item \textit{Is there a brush in the scene?}
        \item \textit{Is there an oven?}
        \item \textit{Is the brush on the oven?}
    \end{enumerate}
\end{enumerate}

\subsection{DPO Training Data Composition Ablation}\label{apx:dpo ablation}
In fine-tuning VLMs to generate questions, we included questions generated through the in-context learning approach introduced in Section~\ref{sec:pmd icl details} to build upon our previous best result (i.e., augmenting VLMs with coherence-based ranking and in-context learning). In Table~\ref{tab:generation metrics apx}, we include an additional result for DPO where generated training data does not include candidate questions generated with in-context learning. As shown, compared to the results in Table~\ref{tab:generation metrics}, the best values of each metric are largely comparable with or without in-context learning, suggesting that in-context learning did not contribute significantly to the performance.

\begin{table}
    \centering
    \setlength{\tabcolsep}{3.5pt}

    \textbf{LLaVA + DPO (without ICL)}

    \vspace{3pt}

    \footnotesize
    \begin{tabular}{ccccccc}\toprule
        \textbf{Rank} & \textbf{ICL} & \textbf{Acc.} $\uparrow$ & \textbf{Rel.} $\uparrow$ & \textbf{Inf.} $\uparrow$ & \textbf{\# Iter.} $\downarrow$ & \textbf{I. Gain} $\uparrow$ \\\cmidrule(lr){1-2}\cmidrule(lr){3-3}\cmidrule(lr){4-5}\cmidrule(lr){6-7}
        L & \xmark & 61.0 & 84.3 & .312 & 2.16 & .746 \\
        L & \cmark & \textbf{64.8} & 64.9 & .326 & 2.76 & .622 \\
        C & \xmark & 61.2 & \textbf{94.6} & .316 & \textbf{1.78} & .772 \\  
        C & \cmark & 61.8 & 93.0 & \textbf{.342} & 2.37 & \textbf{.790} \\\bottomrule
    \end{tabular}

    \normalsize

    \caption{Ego4D-PMD test set results for DPO-trained VLMs without applying in-context learning in generating training data. Inference applies likelihood (L) or coherence (C) candidate question ranking approaches, with optional supplementary candidates generated through in-context learning (ICL).}

    \vspace{-10pt}
    
    \label{tab:generation metrics apx}
\end{table}

\subsection{VLM Self-Dialog Runtime Analysis}\label{apx:timing analysis}

As discussed in Section~\ref{sec:limitations}, as task guidance and PMD will ultimately operate in real-world applications, their responsiveness becomes crucial. While detailed inquiries and methods for more efficient PMD are out of scope for this work, we analyze the runtime of various configurations of LLaVA in Table~\ref{tab:runtime} as information for future work in this area. As expected, coherence-based ranking and in-context learning significantly increase the runtime of models up to about 1 minute per example (not suitable for practical application). Meanwhile, applying the coherence-trained VQG adapter does not have much runtime overhead compared to off-the-shelf LLaVA; both take approximately 3.3 seconds per iteration, and the adapter slightly reduces the total runtime due to taking fewer iterations (more appropriate for practical application).

\begin{table*}
    \centering
    \setlength{\tabcolsep}{2.5pt}

    \footnotesize
    \begin{tabular}{cccccccc}\toprule
        \textbf{Rank} & \textbf{ICL} & \textbf{DPO} & \textbf{Mean \# Iterations} & \textbf{Std. \# Iterations} & \textbf{Mean Runtime} & \textbf{Std. Runtime} & \textbf{Mean Runtime/Iteration} \\\cmidrule(lr){1-3}\cmidrule(lr){4-5}\cmidrule(lr){6-7}\cmidrule(lr){8-8}
        Likelihood & \xmark & \xmark & 3.31 & 1.56 & 11.1 & 5.79 & 3.30 \\
        Coherence & \xmark & \xmark & 3.03 & 1.71 & 30.5 & 20.8 & 9.48 \\
        Coherence & \cmark & \xmark & 3.50 & 2.66 & 61.3 & 53.6 & 16.3 \\
        Coherence & \cmark & \cmark & 1.81 & 1.61 & 28.2 & 28.5 & 14.9 \\
        Likelihood & \xmark & \cmark & 2.26 & 1.37 & 7.74 & 4.92 & 3.38 \\\bottomrule
    \end{tabular}

    \normalsize

    \caption{Runtime analysis for various configurations of LLaVA without and with coherence-based ranking, in-context learning, and a DPO-trained adapter for coherent question generation. Runtimes are measured in seconds per example or iteration. Models are evaluated on the entire validation set of Ego4D-PMD). To better represent a real-time application, models are evaluated one example at a time rather than in batch, which causes small discrepancies with the number of iterations reported in other validation results.}

    \vspace{-10pt}
    
    \label{tab:runtime}
\end{table*}

\subsection{Question Generation Fine-Tuning and Inference Details}\label{apx:dpo details}
When generating the training data from a specific self-dialog iteration, data is omitted if only one candidate question was generated, or if the chosen question has an unsure answer from the VLM (based on the sureness threshold of 60\%). To maximize training data quality, the inference hyperparameters for training data generation are selected based on the training data using the procedure described in Appendix~\ref{apx:selection hyperparam}.
The VQG adapter is trained for 10 epochs with a batch size maximized for our available GPU memory (4). 
The learning rate $\eta$ and DPO $\beta$ hyperparameters are selected from a grid search over combinations of $\eta \in \{ \text{1e-6}, \text{2.5e-6}, \text{5e-6}, \text{7.5e-6}, \text{1e-5} \}$ and $\beta \in \{ 0.05, 0.1, 0.5 \}$, minimizing the minimum validation set DPO loss (as defined by \citealp{rafailov2023direct}) achieved across all epochs. 
In each run, the learning rate is warmed up to its assigned value for the first 5\% of training steps, then linearly decreased to zero through the remaining steps.
Selected hyperparameters for training data generation are listed in Table~\ref{tab:dpo hyperparams training data}, while selected hyperparameters for training and inference are listed in Table~\ref{tab:dpo hyperparams}.
Training is distributed across 4 A40 GPUs, and takes up to about 12 hours. The full validation set results for the results in Tables~\ref{tab:generation metrics}, \ref{tab:generation metrics apx}, and \ref{tab:generation metrics apx 2} with corresponding selected hyperparameters are listed in Table~\ref{tab:generation metrics val}. Note that testing results for DPO without in-context learning in training data generation are introduced earlier in Appendix~\ref{apx:dpo ablation}, while those for DPO with length penalty in training data generation are introduced later in Appendix~\ref{apx:dpo ablation 2}.

\begin{table}
    \centering

    \textbf{LLaVA + DPO}

    \vspace{1pt}

    \footnotesize
    \begin{tabular}{ccc}\toprule
        \textbf{$n^*$} & \textbf{$\delta$} & \textbf{$\epsilon$} \\\cmidrule(lr){1-3} 
        10 & 0.05 & 0.05  \\\bottomrule
    \end{tabular}

    \vspace{6pt}

    \normalsize
    \textbf{LLaVA + DPO (without ICL)}

    \vspace{1pt}

    \footnotesize
    \begin{tabular}{ccc}\toprule
        \textbf{$n^*$} & \textbf{$\delta$} & \textbf{$\epsilon$} \\\cmidrule(lr){1-3} 
        10 & 0.05 & 0.05  \\\bottomrule
    \end{tabular}

    \vspace{6pt}

    \normalsize
    \textbf{LLaVA + DPO (with length penalty)}

    \vspace{1pt}

    \footnotesize
    \begin{tabular}{ccc}\toprule
        \textbf{$n^*$} & \textbf{$\delta$} & \textbf{$\epsilon$} \\\cmidrule(lr){1-3} 
        10 & 0.05 & 0.05  \\\bottomrule
    \end{tabular}

    \normalsize

    \caption{Selected training data generation hyperparameters for the results presented in Tables~\ref{tab:generation metrics}, \ref{tab:generation metrics apx}, and \ref{tab:generation metrics apx 2}.}

    \vspace{-10pt}
    
    \label{tab:dpo hyperparams training data}
\end{table}

\begin{table}
    \centering

    \textbf{LLaVA + DPO}

    \vspace{3pt}

    \footnotesize
    \begin{tabular}{cccccccc}\toprule
        \textbf{Rank} & \textbf{ICL} & \textbf{$\eta$} & \textbf{$\beta$} & \textbf{$n^*$} & \textbf{$\delta$} & \textbf{$\epsilon$} & \textbf{$\tau$} \\\cmidrule(lr){1-2}\cmidrule(lr){3-4}\cmidrule(lr){5-8}   
        L & \xmark & \multirow{4}{*}{1e-5} & \multirow{4}{*}{0.5} & 10 & 0.2 & 0.05 & 0.78 \\
        L & \cmark &  & & 10 & 0.2 & 0.05 & 0.75 \\
        C & \xmark &  & & 10 & 0.2 & 0.05 & 0.58 \\
        C & \cmark &  & & 10 & 0.1 & 0.1 & 0.76 \\\bottomrule
    \end{tabular}

    \vspace{6pt}

    \normalsize
    \textbf{LLaVA + DPO (without ICL)}

    \vspace{1pt}

    \footnotesize
    \begin{tabular}{cccccccc}\toprule
        \textbf{Rank} & \textbf{ICL} & \textbf{$\eta$} & \textbf{$\beta$} & \textbf{$n^*$} & \textbf{$\delta$} & \textbf{$\epsilon$} & \textbf{$\tau$} \\\cmidrule(lr){1-2}\cmidrule(lr){3-4}\cmidrule(lr){5-8}   
        L & \xmark & \multirow{4}{*}{7.5e-6} & \multirow{4}{*}{0.5} & 10 & 0.1 & 0.05 & 0.59 \\
        L & \cmark &  &  & 10 & 0.1 & 0.05 & 0.85 \\
        C & \xmark &  &  & 10 & 0.4 & 0.05 & 0.66 \\
        C & \cmark &  &  & 10 & 0.05 & 0.05 & 0.64 \\\bottomrule
    \end{tabular}

    \vspace{6pt}

    \normalsize
    \textbf{LLaVA + DPO (with length penalty)}

    \vspace{1pt}

    \footnotesize
    \begin{tabular}{cccccccc}\toprule
        \textbf{Rank} & \textbf{ICL} & \textbf{$\eta$} & \textbf{$\beta$} & \textbf{$n^*$} & \textbf{$\delta$} & \textbf{$\epsilon$} & \textbf{$\tau$} \\\cmidrule(lr){1-2}\cmidrule(lr){3-4}\cmidrule(lr){5-8}   
        L & \xmark & \multirow{4}{*}{7.5e-6} &  \multirow{4}{*}{0.1} & 10 & 0.4 & 0.05 & 0.35 \\
        L & \cmark &  &  & 10 & 0.05 & 0.05 & 0.63 \\
        C & \xmark &  &  & 10 & 0.4 & 0.05 & 0.86 \\
        C & \cmark &  &  & 10 & 0.4 & 0.05 & 0.84 \\\bottomrule
    \end{tabular}
    \normalsize

    \caption{Selected training and inference hyperparameters for the results presented in Tables~\ref{tab:generation metrics}, \ref{tab:generation metrics apx}, and \ref{tab:generation metrics apx 2}.}

    \vspace{-10pt}
    
    \label{tab:dpo hyperparams}
\end{table}

\begin{table}
    \centering
    \setlength{\tabcolsep}{3.5pt}

    \textbf{LLaVA + DPO}

    \vspace{3pt}

    \footnotesize
    \begin{tabular}{ccccccc}\toprule
        \textbf{Rank} & \textbf{ICL} & \textbf{Acc.} $\uparrow$ & \textbf{Rel.} $\uparrow$ & \textbf{Inf.} $\uparrow$ & \textbf{\# Iter.} $\downarrow$ & \textbf{I. Gain} $\uparrow$ \\\cmidrule(lr){1-2}\cmidrule(lr){3-3}\cmidrule(lr){4-5}\cmidrule(lr){6-7}
        L & \xmark & 63.0 & 76.7 & .328 & 2.27 & .617 \\
        L & \cmark & 64.0 & 65.5 & \textbf{.347} & 2.48 & .579 \\
        C & \xmark & 63.0 & 91.2 & .342 & 2.11 & .703 \\
        C & \cmark & \textbf{65.2} & \textbf{94.8} & .318 & \textbf{1.85} & \textbf{.727} \\\bottomrule
    \end{tabular}

    \vspace{6pt}

    \normalsize
    \textbf{LLaVA + DPO (without ICL)}

    \vspace{1pt}

    \footnotesize
    \begin{tabular}{ccccccc}\toprule
        \textbf{Rank} & \textbf{ICL} & \textbf{Acc.} $\uparrow$ & \textbf{Rel.} $\uparrow$ & \textbf{Inf.} $\uparrow$ & \textbf{\# Iter.} $\downarrow$ & \textbf{I. Gain} $\uparrow$ \\\cmidrule(lr){1-2}\cmidrule(lr){3-3}\cmidrule(lr){4-5}\cmidrule(lr){6-7}        
        L & \xmark & 62.0 & 85.0 & .293 & 2.04 & .757 \\
        L & \cmark & 62.6 & 71.9 & .291 & 2.40 & .689 \\
        C & \xmark & 62.6 & \textbf{93.6} & .338 & \textbf{1.85} & .756 \\
        C & \cmark & \textbf{63.0} & 92.0 & \textbf{.357} & 2.54 & \textbf{.783} \\\bottomrule
    \end{tabular}

    \vspace{6pt}

    \normalsize
    \textbf{LLaVA + DPO (with length penalty)}

    \vspace{1pt}

    \footnotesize
    \begin{tabular}{ccccccc}\toprule
        \textbf{Rank} & \textbf{ICL} & \textbf{Acc.} $\uparrow$ & \textbf{Rel.} $\uparrow$ & \textbf{Inf.} $\uparrow$ & \textbf{\# Iter.} $\downarrow$ & \textbf{I. Gain} $\uparrow$ \\\cmidrule(lr){1-2}\cmidrule(lr){3-3}\cmidrule(lr){4-5}\cmidrule(lr){6-7}        
        L & \xmark & 62.0 & 93.1 & .293 & 1.97 & .739 \\
        L & \cmark & \textbf{65.4} & 66.1 & \textbf{.412} & 3.61 & .648 \\
        C & \xmark & 61.6 & 97.1 & .281 & 1.86 & .756 \\
        C & \cmark & 61.8 & \textbf{97.2} & .290 & \textbf{1.85} & \textbf{.758} \\\bottomrule
    \end{tabular}

    \normalsize

    \caption{Ego4D-PMD validation set results for DPO-trained VLMs both with and without applying in-context learning in generating training data, as well as with a length penalty applied in generating training data. Inference applies likelihood (L) or coherence (C) candidate question ranking approaches, with optional supplementary candidates generated through in-context learning (ICL).}

    \vspace{-10pt}
    
    \label{tab:generation metrics val}
\end{table}

\subsection{GPT-4o Evaluation}\label{apx:gpt results}
To better contextualize our results with state-of-the-art proprietary large LMs, we additionally evaluated GPT-4o~\cite{openai2024gpt4ocard}\footnote{Specifically, we use the August 6, 2024 version of GPT-4o available in Azure AI Foundry.} on our Ego4D-PMD dataset for coherent PMD. While proprietary models like GPT-4o offer limited customization, making many of the experiments we presented for open-source VLMs, evaluating vanilla GPT-4o serves as a reference point. It is worth noting that even accessed through APIs, GPT-4o returns responses too slowly to be viable for online, frame-by-frame use. 

To enable the GPT-4o evaluation, we make a few small changes to the VLM self-dialog implementation used in the main experiments. Since GPT can not be forced to generate yes-no questions with specific generation constraints, as done in our open-source implementation, we slightly modify the VQG prompt to encourage appropriate question generation. We add the following to the end of our original VQG prompt, with \textbf{bold} indicating the new text:
\begin{quote}

    \textit{This is...from the photo. \textbf{Generate an appropriate yes/no question.}}
    
    \textit{Q:}
\end{quote}

To produce a probability distribution over the \textit{Yes} and \textit{No} tokens, used in both the VQA and success classification parts of the VLM self-dialog, we use the log probabilities provided by the API as logits are not directly available. We then normalize the probabilities to get the final distribution. Since the API only provides the log probabilities of the 20 most likely tokens at every token position, if one of \textit{Yes} and \textit{No} tokens do not appear, we consider its probability to be 0. If both do not appear, we consider their probabilities to be 0.5 each. We also note that we only consider the log probabilities of the token in the first position, as we expect a yes-no answer from GPT. To further encourage this, we make a small addition to the VQA prompt, concatenating ``(yes/no)'' to the end of the question.

We also ran into a couple issues with the Azure OpenAI API. Some specific examples in our dataset triggered Azure's content filter, even when setting the filter's threshold to high. 
In the cases where the issue persisted, we were forced to skip over the example during evaluation. Also, a small portion of API responses returned no completion content during VQG or question rephrasing for NLI (i.e. the \textit{content} attribute of the response had a value of \textit{None}). To deal with this, we implemented a retry procedure, where if no content is given by GPT, we prompt it again with the same prompt. If this second chance also provides no content, we either ignore the example during evaluation in the case of VQG, or simply concatenate the question and answer for question rephrasing. On the validation data, 28 out of 500 examples were omitted, and 103 question candidates could not be rephrased by GPT-4o.

The inference hyperparameters $n^*$, $\delta$, $\epsilon$, and $\tau$ are selected as in our open-source model results, and listed in Table~\ref{tab:gpt hyperparams}. The results on the validation data are listed in Table~\ref{tab:gpt metrics}, while the results on the testing data are listed in the main body of the paper in Table~\ref{tab:selection metrics}. Comparing GPT-4o to the base VLMs we evaluated, it is generally inferior in PMD accuracy and informativeness, but asks more relevant questions, runs for fewer iterations, and has higher information gain. Our best model configurations, though, outperform GPT-4o under all evaluation metrics except information gain.\footnote{LLaVA with coherence-based fine-tuning, coherence-based re-ranking, and in-context learning achieves .742 bits of information gain, while GPT-4o achieves up to .793 bits. This shows that GPT-4o generally exhibits higher confidence despite having much lower PMD accuracy than our approaches, which is not necessarily an advantage.} This suggests that GPT-4o is a reasonable starting point for coherent PMD, but like other off-the-shelf VLMs we evaluated, it may require additional interventions (e.g., coherence-based ranking) to be viable for this task.

\begin{table}
    \centering

    \footnotesize
    \begin{tabular}{cccc}\toprule
        \textbf{$n^*$} & \textbf{$\delta$} & \textbf{$\epsilon$} & \textbf{$\tau$} \\\cmidrule(lr){1-4}  
        10 & 0.4 & 0.1 & 0.19 \\\bottomrule
    \end{tabular}

    \caption{Selected inference hyperparameters for GPT-4o.}

    \vspace{-10pt}
    
    \label{tab:gpt hyperparams}
\end{table}

\begin{table}
    \centering
    \setlength{\tabcolsep}{3.5pt}

    \footnotesize
    \begin{tabular}{ccccc}\toprule
        \textbf{Acc.} $\uparrow$ & \textbf{Rel.} $\uparrow$ & \textbf{Inf.} $\uparrow$ & \textbf{\# Iter.} $\downarrow$ & \textbf{I. Gain} $\uparrow$ \\\cmidrule(lr){1-1}\cmidrule(lr){2-3}\cmidrule(lr){4-5}        
        58.7 & 54.3 & .175 & 1.83 & .730 \\\bottomrule
    \end{tabular}
    \normalsize

    \caption{Ego4D-PMD validation set results for GPT-4o.}

    \vspace{-10pt}
    
    \label{tab:gpt metrics}
\end{table}

\subsection{Rationale-Free Evaluation}\label{apx:exp free results}
For a reference point to incomparable prior works that have applied VLMs to PMD with a focus on classification accuracy \cite{pmlr-v232-du23b,bao-etal-2023-foundation}, we additionally evaluate all studied VLMs from this work on a non-coherent PMD task. Specifically, we prompt each VLM with the following text:

\begin{quote}
  \textit{This is a photo of someone working on the procedure ``$\langle$procedural text$\rangle$''. Q: Based on the image, has the procedure ``$\langle$procedural text$\rangle$'' been successfully completed? A: }
\end{quote}

This prompt is as comparable as possible to the one used for coherent PMD, but does not elicit a series of questions and answers from the VLM. We perform this evaluation on InstructBLIP, LLaVA, Llama 3, and GPT-4o. Table~\ref{tab:exp free hyperparams} lists the inference hyperparameters for this approach, while Table~\ref{tab:exp free metrics} lists the results for the validation and testing data. 
It is crucial to note that \textit{these results are not directly comparable to the coherent PMD results presented in the main body of the paper}, as VLMs are not required to justify their decisions, removing the explainability enabled by coherent PMD (crucial for end users to interpret often incorrect VLM decisions and act on them accordingly). As prior work has already explored this setting more extensively, we do not intend to provide a rigorous study here, rather a reference point to compare how requiring the generation of rationales impacts PMD accuracy.

Interestingly, however, the rationale-free approach achieves generally better accuracy than the coherent PMD results in Tables~\ref{tab:selection metrics} and \ref{tab:selection metrics val} with likelihood-based ranking and no in-context learning. However, upon introducing coherence-based ranking and in-context learning, InstructBLIP and LLaVA achieve better accuracy than in the rationale-free approach. Further, the information gain in the rationale-free setting is consistently lower than those achieved in coherent PMD. This demonstrates that while the added transparency of incorporating rationales into PMD does cost the VLM some accuracy, improving the coherence of these rationales (e.g., through the approaches presented in this work) can recover this accuracy and more while enabling more confident decisions from VLMs. Rationale-free GPT-4o achieves the highest observed accuracy of 69.2\%. Nonetheless, this accuracy is still low enough for errors to be common, thus necessitating the generation of a rationale for the user.

\begin{table}
    \centering















    \footnotesize
    \begin{tabular}{cccc}\toprule
        \textbf{InstructBLIP} & \textbf{LLaVA} & \textbf{Llama 3} & \textbf{GPT-4o} \\\cmidrule(lr){1-4}  
        0.34 & 0.77 & 0.54 & 0.98 \\\bottomrule
    \end{tabular}    

    \caption{Selected values of mistake confidence threshold $\tau$ for rationale-free PMD with various VLMs. $n^*$, $\delta$, and $\epsilon$ are no longer used in rationale-free PMD, but $\tau$ is still tuned as in previous experiments.}

    \vspace{-10pt}
    
    \label{tab:exp free hyperparams}
\end{table}

\begin{table}
    \centering
    \setlength{\tabcolsep}{3.5pt}

    \vspace{6pt}

    \normalsize
    \textbf{InstructBLIP}

    \vspace{1pt}

    \footnotesize
    \begin{tabular}{cccccc}\toprule
        \textbf{Partition} & \textbf{Acc.} $\uparrow$ & \textbf{Rel.} $\uparrow$ & \textbf{Inf.} $\uparrow$ & \textbf{\# Iter.} $\downarrow$ & \textbf{I. Gain} $\uparrow$ \\\cmidrule(lr){1-1}\cmidrule(lr){2-2}\cmidrule(lr){3-4}\cmidrule(lr){5-6}        
        Validation  & 62.6 & -- & -- & 0.00 & .113 \\\cmidrule(lr){1-1}\cmidrule(lr){2-2}\cmidrule(lr){3-4}\cmidrule(lr){5-6}  
        Testing  & 62.2 & -- & -- & 0.00 & .117 \\\bottomrule
    \end{tabular}

    \vspace{6pt}

    \normalsize
    \textbf{LLaVA}

    \vspace{1pt}

    \footnotesize
    \begin{tabular}{cccccc}\toprule
        \textbf{Partition} & \textbf{Acc.} $\uparrow$ & \textbf{Rel.} $\uparrow$ & \textbf{Inf.} $\uparrow$ & \textbf{\# Iter.} $\downarrow$ & \textbf{I. Gain} $\uparrow$ \\\cmidrule(lr){1-1}\cmidrule(lr){2-2}\cmidrule(lr){3-4}\cmidrule(lr){5-6}        
        Validation  & 64.4 & -- & -- & 0.00 & .236 \\\cmidrule(lr){1-1}\cmidrule(lr){2-2}\cmidrule(lr){3-4}\cmidrule(lr){5-6}  
        Testing  & 66.1 & -- & -- & 0.00 & .233 \\\bottomrule
    \end{tabular}

    \vspace{6pt}

    \normalsize
    \textbf{Llama 3}

    \vspace{1pt}

    \footnotesize
    \begin{tabular}{cccccc}\toprule
        \textbf{Partition} & \textbf{Acc.} $\uparrow$ & \textbf{Rel.} $\uparrow$ & \textbf{Inf.} $\uparrow$ & \textbf{\# Iter.} $\downarrow$ & \textbf{I. Gain} $\uparrow$ \\\cmidrule(lr){1-1}\cmidrule(lr){2-2}\cmidrule(lr){3-4}\cmidrule(lr){5-6}        
        Validation  & 65.8 & -- & -- & 0.00 & .187 \\\cmidrule(lr){1-1}\cmidrule(lr){2-2}\cmidrule(lr){3-4}\cmidrule(lr){5-6}  
        Testing  & 64.6 & -- & -- & 0.00 & .176 \\\bottomrule
    \end{tabular}

    \vspace{6pt}

    \normalsize
    \textbf{GPT-4o}

    \vspace{1pt}

    \footnotesize
    \begin{tabular}{cccccc}\toprule
        \textbf{Partition} & \textbf{Acc.} $\uparrow$ & \textbf{Rel.} $\uparrow$ & \textbf{Inf.} $\uparrow$ & \textbf{\# Iter.} $\downarrow$ & \textbf{I. Gain} $\uparrow$ \\\cmidrule(lr){1-1}\cmidrule(lr){2-2}\cmidrule(lr){3-4}\cmidrule(lr){5-6}        
        Validation  & 66.8 & -- & -- & 0.00 & .718 \\\cmidrule(lr){1-1}\cmidrule(lr){2-2}\cmidrule(lr){3-4}\cmidrule(lr){5-6}  
        Testing  & 69.2 & -- & -- & 0.00 & .736  \\\bottomrule
    \end{tabular}    
    
    \normalsize

    \caption{Ego4D-PMD validation and testing set results for rationale-free PMD with various VLMs. As self-dialog rationales are no longer generated, relevance and informativeness cannot be calculated. Further, zero iterations are performed, and information gain is calculated using the average entropy of the success probability without any rationale.}

    \vspace{-10pt}
    
    \label{tab:exp free metrics}
\end{table}

\subsection{Question Selection Inference Hyperparameters and Validation Results}\label{apx:selection hyperparam}
\begin{figure*}
    \centering
    \begin{multicols}{3}
    
    \textbf{InstructBLIP}
    
    \includegraphics[width=1.0\linewidth]{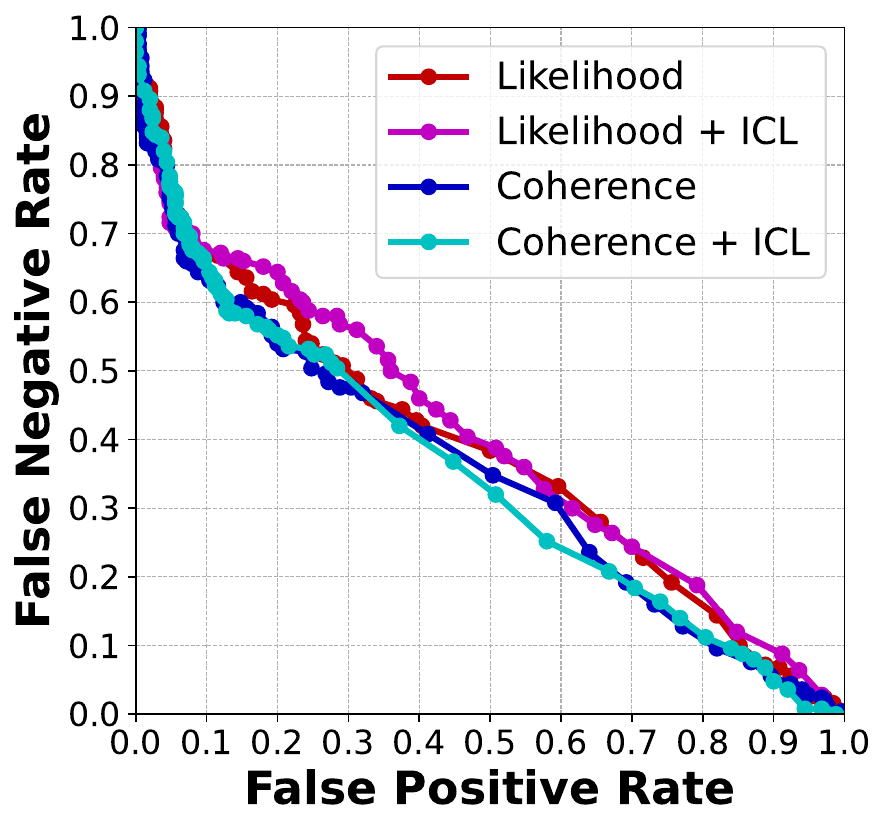}

    \textbf{LLaVA}
    
    \includegraphics[width=1.0\linewidth]{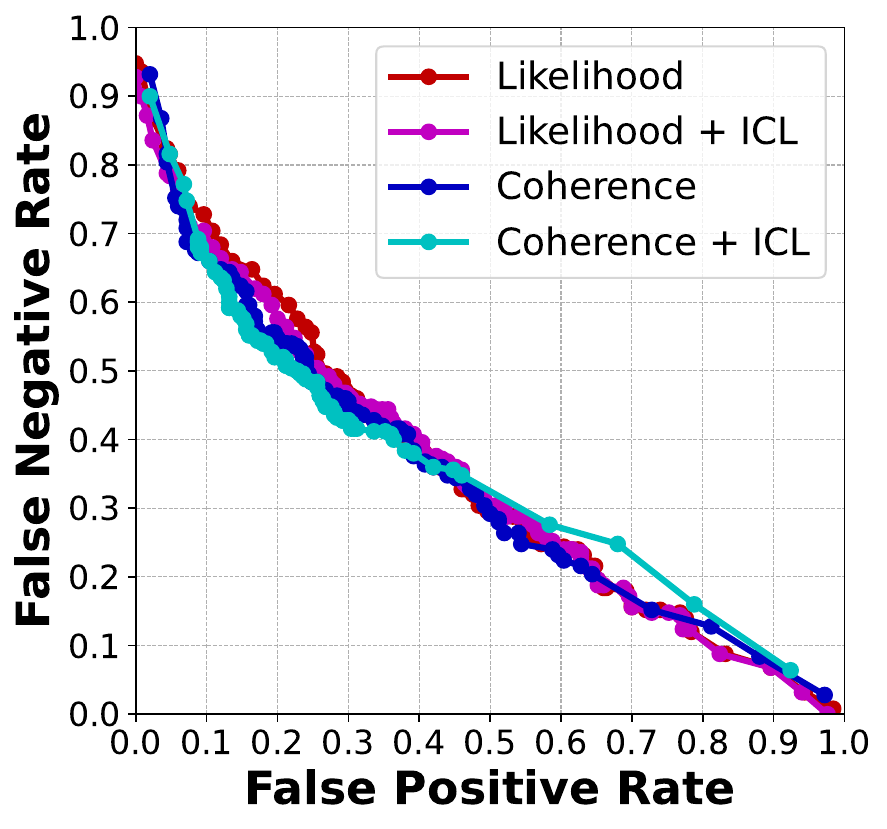}

    \textbf{Llama 3}
    
    \includegraphics[width=1.0\linewidth]{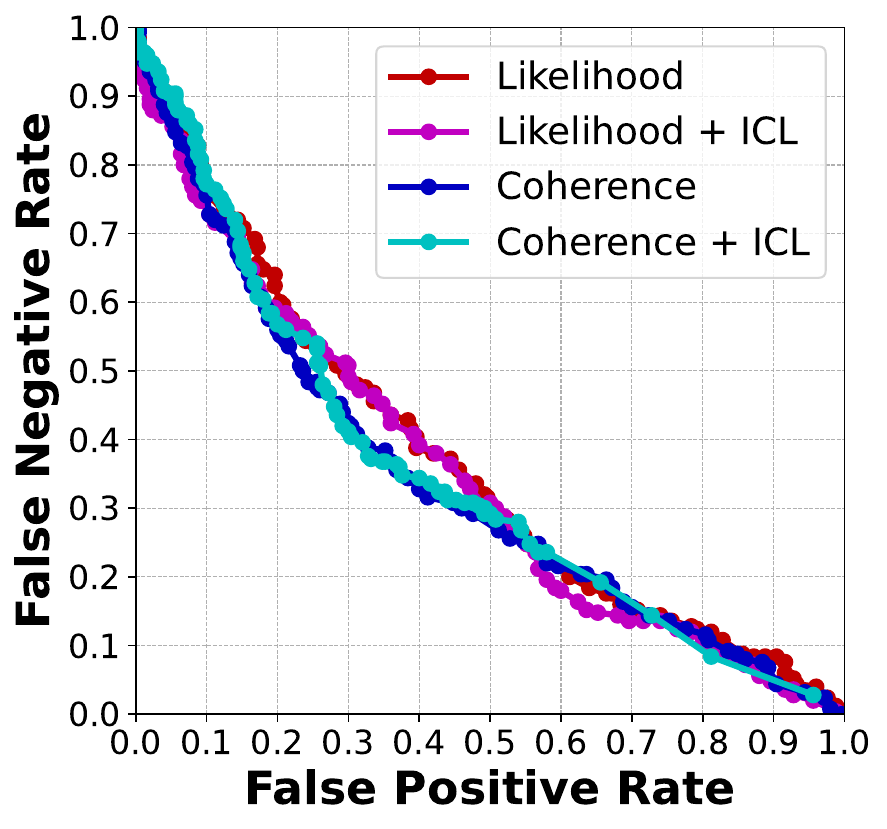}

    \end{multicols}

    \vspace{-15pt}
    \caption{Mistake detection error tradeoff (DET) curves for VLMs applied to the Ego4D-PMD validation set with likelihood- and coherence-based candidate question selection approaches, with optional supplementary candidates generated through in-context learning (ICL).}
    \label{fig:selection det}
\end{figure*}

In Table~\ref{tab:selection hyperparams}, we list the inference hyperparameters for the question selection results presented in Table~\ref{tab:selection metrics}: maximum number of iterations $n^*$, early stopping $\delta$ and $\epsilon$, and mistake confidence threshold $\tau$. 
In Figure~\ref{fig:selection det}, we use detection error tradeoff (DET) curves to visualize the range of accuracy achieved with all candidate mistake confidence thresholds $\tau$ for the approaches compared in Table~\ref{tab:selection metrics}.
The full validation set results with selected hyperparameters are listed in Table~\ref{tab:selection metrics val}.

\begin{table}
    \centering

    \textbf{InstructBLIP}

    \vspace{1pt}

    \footnotesize
    \begin{tabular}{cccccc}\toprule
        \textbf{Rank} & \textbf{ICL} & \textbf{$n^*$} & \textbf{$\delta$} & \textbf{$\epsilon$} & \textbf{$\tau$} \\\cmidrule(lr){1-2}\cmidrule(lr){3-6} 
        L & \xmark & 10 & 0.1 & 0.2 & 0.41 \\
        L & \cmark & 10 & 0.05 & 0.1 & 0.49 \\
        C & \xmark & 10 & 0.05 & 0.2 & 0.35 \\
        C & \cmark & 10 & 0.05 & 0.2 & 0.33 \\\bottomrule
    \end{tabular}

    \vspace{6pt}

    \normalsize
    \textbf{LLaVA}

    \vspace{1pt}

    \footnotesize
    \begin{tabular}{cccccc}\toprule
    \textbf{Rank} & \textbf{ICL} & \textbf{$n^*$} & \textbf{$\delta$} & \textbf{$\epsilon$} & \textbf{$\tau$} \\\cmidrule(lr){1-2}\cmidrule(lr){3-6} 
        L & \xmark & 10 & 0.1 & 0.05 & 0.64 \\
        L & \cmark & 10 & 0.1 & 0.05 & 0.72 \\
        C & \xmark & 10 & 0.1 & 0.05 & 0.76 \\
        C & \cmark & 10 & 0.05 & 0.05 & 0.74 \\\bottomrule
    \end{tabular}

    \vspace{6pt}

    \normalsize
    \textbf{Llama 3}

    \vspace{1pt}

    \footnotesize
    \begin{tabular}{cccccc}\toprule
    \textbf{Rank} & \textbf{ICL} & \textbf{$n^*$} & \textbf{$\delta$} & \textbf{$\epsilon$} & \textbf{$\tau$}  \\\cmidrule(lr){1-2}\cmidrule(lr){3-6} 
    L & \xmark & 10 & 0.1 & 0.05 & 0.40 \\
    L & \cmark & 10 & 0.05 & 0.025 & 0.23 \\
    C & \xmark & 10 & 0.05 & 0.05 & 0.38 \\
    C & \cmark & 10 & 0.2 & 0.05 & 0.30 \\\bottomrule
    \end{tabular}

    \normalsize

    \caption{Selected inference hyperparameters for the results presented in Table~\ref{tab:selection metrics}.}

    \vspace{-10pt}
    
    \label{tab:selection hyperparams}
\end{table}

\begin{table}[!ht]
    \centering
    \setlength{\tabcolsep}{3.5pt}

    \textbf{InstructBLIP}

    \vspace{1pt}

    \footnotesize
    \begin{tabular}{ccccccc}\toprule
        \textbf{Rank} & \textbf{ICL} & \textbf{Acc.} $\uparrow$ & \textbf{Rel.} $\uparrow$ & \textbf{Inf.} $\uparrow$ & \textbf{\# Iter.} $\downarrow$ & \textbf{I. Gain} $\uparrow$ \\\cmidrule(lr){1-2}\cmidrule(lr){3-3}\cmidrule(lr){4-5}\cmidrule(lr){6-7}
        L & \xmark & 62.0 & 18.3 & .237 & \textbf{2.79} & .265 \\
        L & \cmark & 61.8 & 14.0 & .325 & 4.62 & .358 \\
        C & \xmark & 63.8 & 26.0 & .285 & 3.52 & .298 \\
        C & \cmark & \textbf{64.2} & \textbf{36.2} & \textbf{.336} & 3.30 & \textbf{.363} \\\bottomrule
    \end{tabular}

    \vspace{6pt}

    \normalsize
    \textbf{LLaVA}

    \vspace{1pt}

    \footnotesize
    \begin{tabular}{ccccccc}\toprule
        \textbf{Rank} & \textbf{ICL} & \textbf{Acc.} $\uparrow$ & \textbf{Rel.} $\uparrow$ & \textbf{Inf.} $\uparrow$ & \textbf{\# Iter.} $\downarrow$ & \textbf{I. Gain} $\uparrow$ \\\cmidrule(lr){1-2}\cmidrule(lr){3-3}\cmidrule(lr){4-5}\cmidrule(lr){6-7}        
        L & \xmark & 62.0 & 42.7 & .287 & 3.15 & .437 \\
        L & \cmark & 62.2 & 41.7 & .289 & 3.17 & .439 \\
        C & \xmark & 63.6 & 68.5 & .319 & \textbf{3.05} & .532 \\
        C & \cmark & \textbf{64.4} & \textbf{76.5} & \textbf{.418} & 3.45 & \textbf{.659} \\\bottomrule
    \end{tabular}

    \vspace{6pt}

    \normalsize
    \textbf{Llama 3}

    \vspace{1pt}

    \footnotesize
    \begin{tabular}{ccccccc}\toprule
        \textbf{Rank} & \textbf{ICL} & \textbf{Acc.} $\uparrow$ & \textbf{Rel.} $\uparrow$ & \textbf{Inf.} $\uparrow$ & \textbf{\# Iter.} $\downarrow$ & \textbf{I. Gain} $\uparrow$ \\\cmidrule(lr){1-2}\cmidrule(lr){3-3}\cmidrule(lr){4-5}\cmidrule(lr){6-7}        
        L & \xmark & 60.8 & 16.8 & .287 & 4.62 & .219 \\
        L & \cmark & 61.2 & 16.2 & .335 & 6.38 & .245 \\
        C & \xmark & 64.2 & 25.0 & .347 & 6.37 & .236 \\
        C & \cmark & \textbf{64.8} & \textbf{52.3} & \textbf{.469} & \textbf{3.57} & \textbf{.396} \\\bottomrule
    \end{tabular}
    \normalsize
    \caption{Ego4D-PMD validation set results for likelihood-based (L) and coherence-based (C) candidate question ranking approaches, with optional supplementary candidates generated through in-context learning (ICL).}

    
    \label{tab:selection metrics val}
\end{table}

\subsection{Analysis of Question Sources in In-Context Learning}\label{apx:icl hist}

\begin{figure}
    \centering

    \textbf{InstructBLIP}

    \includegraphics[width=0.95\linewidth]{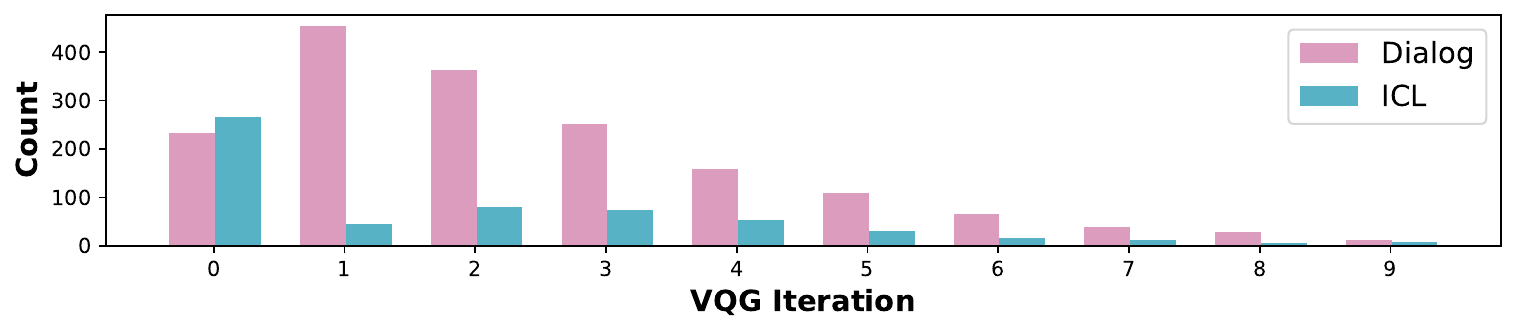} \\
    \includegraphics[width=0.95\linewidth]{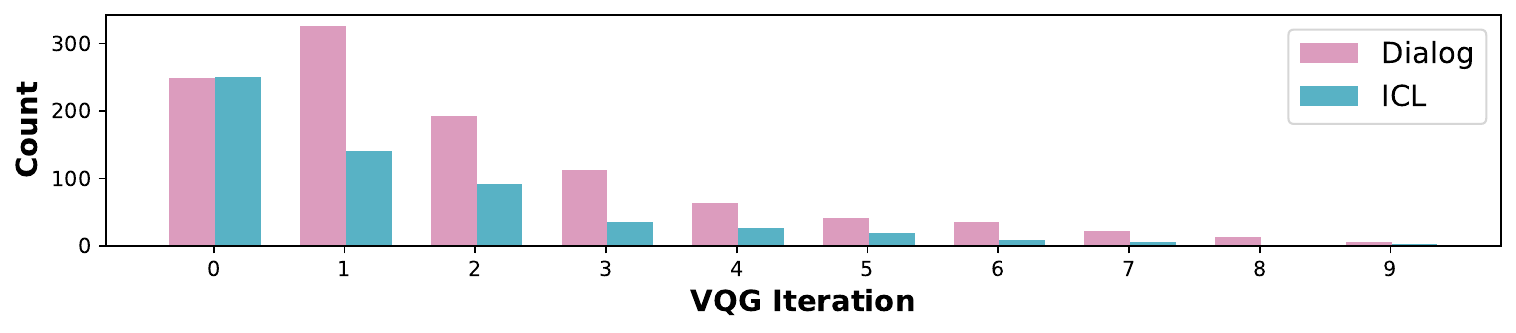}

    \vspace{6pt}
    
    \textbf{LLaVA}
    
    \includegraphics[width=0.95\linewidth]{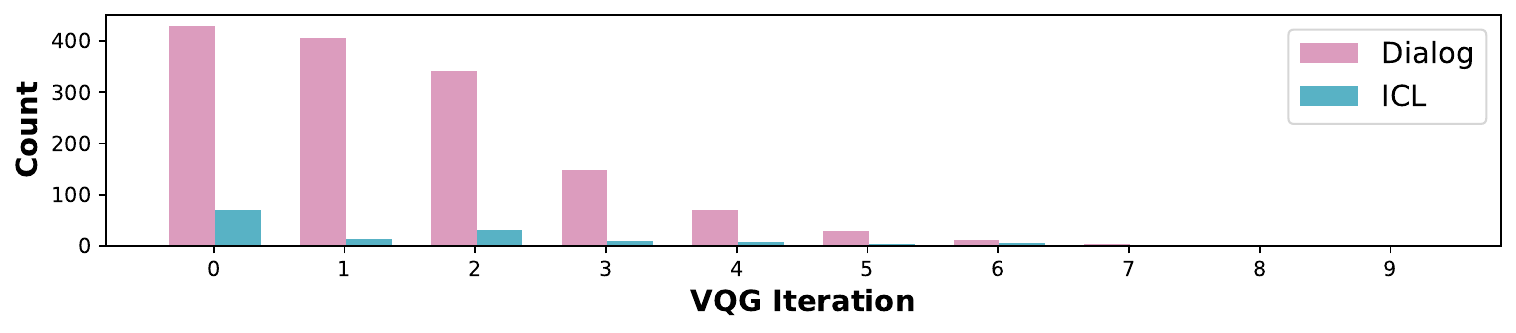} \\
    \includegraphics[width=0.95\linewidth]{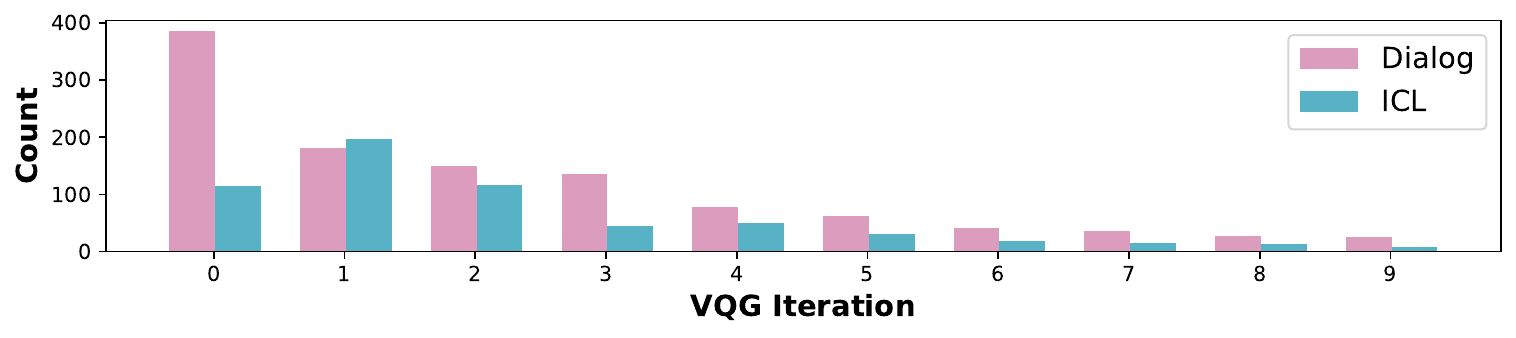}
    
    \vspace{6pt}
    
    \textbf{Llama 3}
    
    \includegraphics[width=0.95\linewidth]{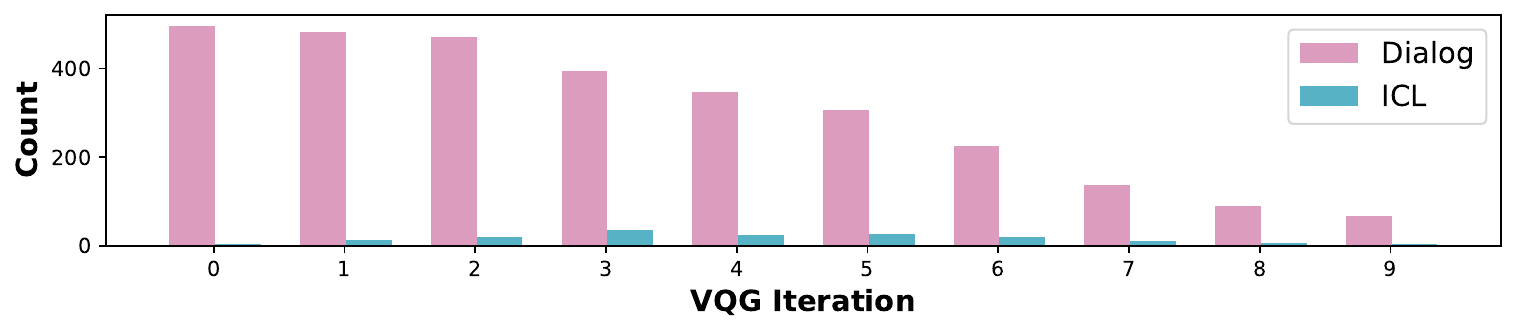} \\
    \includegraphics[width=0.95\linewidth]{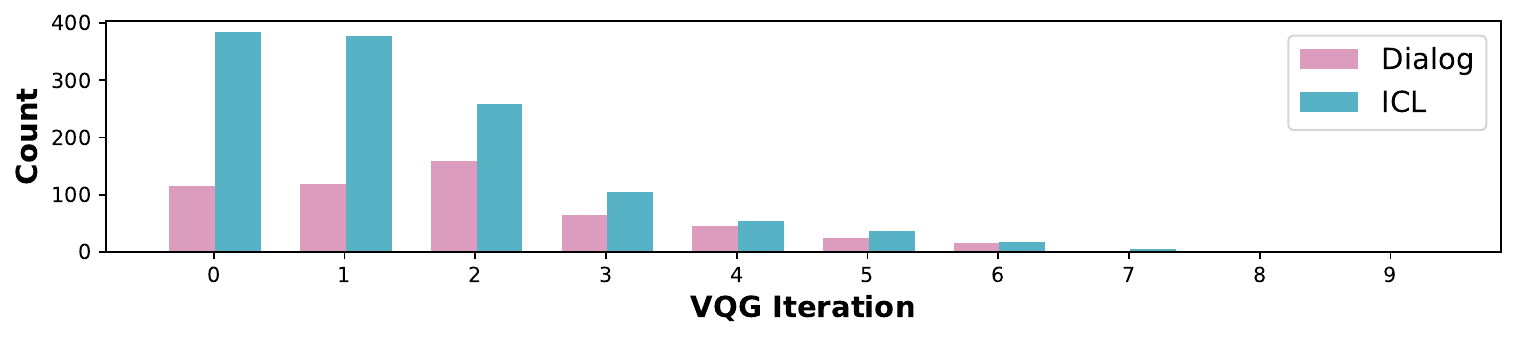}
    
    \caption{Histograms of VLMs' selected question sources, either self-dialog context or in-context learning (ICL) examples, by visual question generation (VQG) iteration for likelihood-based question selection (top) and coherence-based question selection (bottom).}
    \label{fig:icl hist}
\end{figure}

To shed more light on where selected candidate questions come from in each approach, we visualize the distribution of question sources on the validation data in Figure~\ref{fig:icl hist}. As expected, candidates generated with in-context learning are relatively rarely selected under likelihood-based ranking, amounting to about 25.7\% of VQG iterations for InstructBLIP, 9.0\% of VQG iterations for LLaVA, and 5.2\% of VQG iterations for Llama 3. On the other hand, they are selected more frequently in the coherence-based ranking, amounting to about 35.6\% of VQG iterations for InstructBLIP, 30.6\% of VQG iterations for LLaVA, and 36.5\% of VQG iterations for Llama 3. Interestingly, in-context learning candidates are more dominant in earlier iterations, while candidates generated based on the dialog context are relatively more common in later iterations. This may suggest that after selecting a few questions from in-context learning in earlier iterations, the VLM is able to utilize them to generate better questions from the dialog context in later iterations. Alternatively, this could suggest that candidates from in-context learning have limited variety, and thus are less likely to be selected in later turns to avoid redundant questions or information.

\subsection{Diversity-Based Ranking Baseline}\label{apx:distance based}
One possible explanation for the effectiveness of coherence-based question ranking in Section~\ref{sec:question selection} is that it enables the selection of more semantically diverse questions, thus collecting broader information about the image. To explore this question, we implement a supplemental diversity-based ranking approach which uses a sentence transformer~\cite{reimers-2020-multilingual-sentence-bert}\footnote{See \url{https://huggingface.co/sentence-transformers/all-MiniLM-L6-v2}.} to embed all previous and candidate questions at each iteration, then selects the candidate question with the largest average cosine distance from previous questions.

As shown in Table~\ref{tab:diversity metrics}, we observe that diversity-based ranking combined with in-context learning can also improve the accuracy of VLMs (to a level slightly below that of coherence-based ranking). Accuracy reaches respective maxima of 64.7\% and 67.1\% for InstructBLIP and LLaVA (compared to 66.6\% and 67.8\% under coherence-based ranking). This may suggest that some accuracy improvements in coherence-based ranking could be attributed simply to the ability to select more diverse questions than a likelihood-based approach. Additioanlly, this reaffirms our observation that VLMs are poorly suited for this task off-the-shelf.

However, we also see that in most cases, Diversity-based ranking substantially degrades relevance, informativeness, number of iterations, and information gain compared to those in Table~\ref{tab:selection metrics}. This suggests that the highly exploratory nature of diversity-based ranking causes rationales to be less coherent, and conclusions are made slower and less confidently with this approach. Meanwhile, coherence-based ranking enables us to find good questions to ask faster, leading to more confident conclusions with more relevant and informative supporting evidence (while also achieving a higher accuracy).

\begin{table}
    \centering

    \textbf{InstructBLIP}

    \vspace{1pt}

    \footnotesize
    \begin{tabular}{ccccc}\toprule
    \textbf{ICL} & \textbf{$n^*$} & \textbf{$\delta$} & \textbf{$\epsilon$} & \textbf{$\tau$}  \\\cmidrule(lr){1-1}\cmidrule(lr){2-5} 
        \xmark & 10 & 0.1 & 0.2 & 0.25 \\
        \cmark & 10 & 0.1 & 0.1 & 0.23 \\\bottomrule
    \end{tabular}

    \vspace{6pt}

    \normalsize
    \textbf{LLaVA}

    \vspace{1pt}

    \footnotesize
    \begin{tabular}{ccccc}\toprule
    \textbf{ICL} & \textbf{$n^*$} & \textbf{$\delta$} & \textbf{$\epsilon$} & \textbf{$\tau$}  \\\cmidrule(lr){1-1}\cmidrule(lr){2-5} 
        \xmark & 10 & 0.05 & 0.05 & 0.58 \\
        \cmark & 10 & 0.05 & 0.05 & 0.69 \\\bottomrule
    \end{tabular}

    \vspace{6pt}

    \normalsize
    \textbf{Llama 3}

    \vspace{1pt}

    \footnotesize
    \begin{tabular}{ccccc}\toprule
    \textbf{ICL} & \textbf{$n^*$} & \textbf{$\delta$} & \textbf{$\epsilon$} & \textbf{$\tau$}  \\\cmidrule(lr){1-1}\cmidrule(lr){2-5} 
        \xmark & 10 & 0.05 & 0.05 & 0.41 \\
        \cmark & 10 & 0.1 & 0.025 & 0.40 \\\bottomrule
    \end{tabular}

    \normalsize

    \caption{Selected inference hyperparameters for diversity-based ranking with various VLMs, with optional supplementary candidates generated through in-context learning (ICL).}

    \vspace{-10pt}
    
    \label{tab:selection hyperparams diversity}
\end{table}

\begin{table}
    \centering
    \setlength{\tabcolsep}{2.3pt}

    \vspace{6pt}

    \normalsize
    \textbf{InstructBLIP}

    \vspace{1pt}

    \footnotesize
    \begin{tabular}{ccccccc}\toprule
        \textbf{Partition} & \textbf{ICL} & \textbf{Acc.} $\uparrow$ & \textbf{Rel.} $\uparrow$ & \textbf{Inf.} $\uparrow$ & \textbf{\# Iter.} $\downarrow$ & \textbf{I. Gain} $\uparrow$ \\\cmidrule(lr){1-1}\cmidrule(lr){2-2}\cmidrule(lr){3-3}\cmidrule(lr){4-5}\cmidrule(lr){6-7}        
        Validation  & \xmark & 61.4 & \textbf{17.8} & .244 & \textbf{2.89} & .268 \\
        Validation  & \cmark & \textbf{64.0} & 14.4 & \textbf{.337} & 3.82 & \textbf{.350} \\\cmidrule(lr){1-1}\cmidrule(lr){2-2}\cmidrule(lr){3-3}\cmidrule(lr){4-5}\cmidrule(lr){6-7}
        Testing & \xmark & 64.2 & \textbf{17.5} & .237 & \textbf{2.88} & .268  \\
        Testing & \cmark & \textbf{64.7} & 15.1 & \textbf{.337} & 3.72 & \textbf{.344}  \\\bottomrule
    \end{tabular}    

    \vspace{6pt}

    \normalsize
    \textbf{LLaVA}

    \vspace{1pt}

    \footnotesize
    \begin{tabular}{ccccccc}\toprule
        \textbf{Partition} & \textbf{ICL} & \textbf{Acc.} $\uparrow$ & \textbf{Rel.} $\uparrow$ & \textbf{Inf.} $\uparrow$ & \textbf{\# Iter.} $\downarrow$ & \textbf{I. Gain} $\uparrow$ \\\cmidrule(lr){1-1}\cmidrule(lr){2-2}\cmidrule(lr){3-3}\cmidrule(lr){4-5}\cmidrule(lr){6-7}        
        Validation  & \xmark & 64.2 & 37.0 & .298 & 4.12 & .444 \\
        Validation  & \cmark & \textbf{65.2} & \textbf{43.2} & \textbf{.385} & \textbf{3.99} & \textbf{.539} \\\cmidrule(lr){1-1}\cmidrule(lr){2-2}\cmidrule(lr){3-3}\cmidrule(lr){4-5}\cmidrule(lr){6-7}
        Testing & \xmark & 62.9 & 35.2 & .310 & 4.35 & .455  \\
        Testing & \cmark & \textbf{67.1} & \textbf{39.8} & \textbf{.405} & \textbf{4.32} & \textbf{.513}  \\\bottomrule
    \end{tabular}    

    \vspace{6pt}

    \normalsize
    \textbf{Llama 3}

    \vspace{1pt}

    \footnotesize
    \begin{tabular}{ccccccc}\toprule
        \textbf{Partition} & \textbf{ICL} & \textbf{Acc.} $\uparrow$ & \textbf{Rel.} $\uparrow$ & \textbf{Inf.} $\uparrow$ & \textbf{\# Iter.} $\downarrow$ & \textbf{I. Gain} $\uparrow$ \\\cmidrule(lr){1-1}\cmidrule(lr){2-2}\cmidrule(lr){3-3}\cmidrule(lr){4-5}\cmidrule(lr){6-7}        
        Validation  & \xmark & 62.6 & 15.7 & .313 & 6.19 & .262 \\
        Validation  & \cmark & \textbf{66.8} & \textbf{24.2} & \textbf{.412} & \textbf{5.14} & \textbf{.358} \\\cmidrule(lr){1-1}\cmidrule(lr){2-2}\cmidrule(lr){3-3}\cmidrule(lr){4-5}\cmidrule(lr){6-7}
        Testing & \xmark & \textbf{60.9} & 15.1 & .322 & 6.48 & .246  \\
        Testing & \cmark & 59.6 & \textbf{22.6} & \textbf{.393} & \textbf{5.23} & \textbf{.332}  \\\bottomrule
    \end{tabular}    

    \vspace{6pt}
    
    \normalsize

    \caption{Ego4D-PMD validation and testing set results for diversity-based ranking with various VLMs, with optional supplementary candidates generated through in-context learning (ICL).}

    \vspace{-10pt}
    
    \label{tab:diversity metrics}
\end{table}

\subsection{DPO with Length Penalty Experiment}\label{apx:dpo ablation 2}
In the results discussed in Section~\ref{sec: dpo results}, we observed that while VLMs learned to generate much more relevant questions, the informativeness of answers and thus the PMD accuracy degraded. We hypothesized that this resulted from the generation of highly complex questions, e.g., ``Is the soil placed around the seedling with the trowel in the person's hand?'' As such, a potentially fruitful avenue for future research is to explore decoding approaches and learning objectives that prioritize more approachable questions for VLMs. 

While the primary purpose of this work was to recast the problem of PMD into a more transparent formulation and lay a foundation for research toward coherent question generation and answering in PMD, we performed an initial experiment to guide future work along this line. Specifically, we imposed an exponential \textit{length penalty} $l=-1.0$ to text generation during training data generation and inference for evaluation.\footnote{Like the results presented in Table~\ref{tab:generation metrics}, candidates generated through in-context learning are also included in training data.} During beam search, a length penalty modifies the total log-likelihood $p_{Q}$ of a partial candidate question $Q$ as follows:

\begin{equation*}
    p'_{Q} = \frac{p_Q}{{\vert Q \vert }^{l}}
\end{equation*}

Here, $\vert Q \vert$ is the length of the question $Q$, i.e., number of generated tokens thus far. For earlier experiments, the value of $l$ could be thought of as 1, which effectively applies no exponential penalty to $\vert Q \vert$ when calculating likelihood-based sequence scores (the default behavior).

In Table~\ref{tab:generation metrics apx 2}, we list the results of applying this length penalty. As shown, compared to the results in Table~\ref{tab:generation metrics}, similar trends of hold despite applying the length penalty: we observe improved relevance, number of iterations, and information gain, but degraded accuracy and informativeness. An interesting exception is when using likelihood-based ranking and in-context learning during inference, we recover a comparable accuracy and informativeness to those observed in LLaVA before applying DPO, but this comes at a cost of a a lower relevance, higher number of iterations, and lower information gain than other inference configurations. This provides further evidence that there exists a trade-off between generating relevant questions and achieving high informativeness and accuracy, and future work should aim to find a balance between these priorities.

\begin{table}
    \centering
    \setlength{\tabcolsep}{3.5pt}

    \textbf{LLaVA + DPO (with length penalty)}

    \vspace{3pt}

    \footnotesize
    \begin{tabular}{ccccccc}\toprule
        \textbf{Rank} & \textbf{ICL} & \textbf{Acc.} $\uparrow$ & \textbf{Rel.} $\uparrow$ & \textbf{Inf.} $\uparrow$ & \textbf{\# Iter.} $\downarrow$ & \textbf{I. Gain} $\uparrow$ \\\cmidrule(lr){1-2}\cmidrule(lr){3-3}\cmidrule(lr){4-5}\cmidrule(lr){6-7}
        L & \xmark & 61.7 & 93.5 & .293 & 1.93 & .741 \\
        L & \cmark & \textbf{67.0} & 56.4 & \textbf{.459} & 4.05 & .628 \\
        C & \xmark & 63.1 & 97.5 & .313 & \textbf{1.82} & \textbf{.771} \\  
        C & \cmark & 62.9 & \textbf{97.7} & .319 & 1.83 & .769 \\\bottomrule
    \end{tabular}

    \normalsize

    \caption{Ego4D-PMD test set results for DPO-trained VLMs with an additional length penalty $l=-1$ applied during training data generation and inference. Inference applies likelihood (L) or coherence (C) candidate question ranking approaches, with optional supplementary candidates generated through in-context learning (ICL).}

    \vspace{-10pt}
    
    \label{tab:generation metrics apx 2}
\end{table}

\subsection{Using Coherence Metrics to Diagnose Common VLM Behaviors}\label{apx:cube graph examples}

To deepen the insights from the graphs in Figure~\ref{fig:cubes}, in Figure~\ref{fig:cube graph examples}, we provided several example outputs from LLaVA with coherence-based ranking, which displays a range of behaviors. Below, we further explain these behaviors and examples.

\paragraph{Correct and coherent points.}
Cyan points have low error with high informativeness and relevance, indicating correct decisions with coherent rationales. These are the best case examples from the model. Figure~\ref{fig:cube graph examples}, Example A is one such case, where LLaVA correctly determines that the procedure ``Pick up a sink brush from the kitchen slab'' has been successfully completed, rationalizing it coherently and succinctly with a single question and answer about the location of the \textit{sink brush}.

\paragraph{Incorrect and incoherent points.}
Conversely, red to magenta points have high error, low informativeness, and low relevance, indicating incorrect decisions with incoherent rationales. These are the worst case examples from the model. Figure~\ref{fig:cube graph examples}, Example B is one such case, where LLaVA incorrectly decides that the procedure ``Tighten the screw'' was not successfully completed due to the person in the image not wearing various protective gear, an incoherent rationale for the decision.

\paragraph{Correct but incoherent points.}
Indigo to black points have low error, but low relevance and informativeness, indicating correct decisions without sufficient rationale. Figure~\ref{fig:cube graph examples}, Example C, is an instance of this, where LLaVA correctly decides that the person in the image has not successfully completed the procedure ``Paint the stone'' (rather they are painting a wood molding). However, LLaVA's decision is only supported by a question about whether the person is wearing a shirt, which it does not answer confidently, making for a completely insufficient rationale.

\paragraph{Coherent but incorrect points.}
White points have high error, relevance, and informativeness, indicating coherent rationales that do not lead to a correct decision. In other words, the information collected by the VLM should have been sufficient to make a correct decision (according to our automated coherence metrics), but this did not occur. Figure~\ref{fig:cube graph examples}, Example D shows one such case, where LLaVA incorrectly decides that the procedure ``Drop the bottle of mustard on the countertop'' was unsuccessful. While it correctly identified that \textit{the bottle} is on \textit{the countertop}, which suggests the success of the procedure, it later mistakenly identified \textit{the bottle} to be on the floor, creating a contradiction in the rationale and causing it to make the wrong decision. The ability of this analysis to easily identify issues like this may be useful for future work in PMD and task guidance, as it enables the detection and thus the correction of bugs in the system's reasoning.

\paragraph{Irrelevant but informative points.}
Blue points have low relevance but relatively high informativeness, indicating irrelevant questions that still yield informative answers. As shown in Figure~\ref{fig:cube graph examples}, Example E, this does not necessarily indicate a failure of LLaVA, rather a terse rationale. In this example, LLaVA correctly determines that the procedure ``Fold the cut piece of cloth'' has not been completed successfully. It reasonably rationalizes this decision by asking about the presence of a \textit{piece of cloth} and responding with \textit{No}. The question of whether the person is working with a \textit{piece of cloth} is deemed somewhat irrelevant by our metrics because if the answer were instead \textit{Yes}, this would not provide sufficient information to conclude that the procedure was successful. However, since the answer was \textit{No}, we do have sufficient information to conclude that the procedure is unsuccessful, despite the question being relatively indirect. Blue points may thus point to sufficient rationales which lack some detail or specificity, which are not necessarily problematic to system performance.

\paragraph{Relevant but uninformative points.}
Green and yellow points have high relevance but low informativeness, indicating a failure to extract useful information in VQA. Green points have close to zero informativeness, indicating unsure responses in VQA. In Figure~\ref{fig:cube graph examples}, Example F, LLaVA rationalizes its decision about the procedure ``Put the bolt remover in the lawn tractor'' by asking whether \textit{the bolt remover} is in \textit{the lawn tractor} in various ways. However, these objects are not present in the image and thus LLaVA's answer is not confident, causing it to respond \textit{Unsure} to most questions, leading to low informativeness. Despite its failure to answer questions, LLaVA still arrives at the correct conclusion that the procedure has not been successfully completed.

Meanwhile, yellow points have highly negative informativeness, indicating counterproductive responses in VQA that oppose the correct decision. As shown in Figure~\ref{fig:cube graph examples}, Examples G and H, these cases typically occur when the VLM does not recognize an object in the image, or it recognizes an object that is not in the image. In Example G, LLaVA incorrectly decides that the procedure ``Put the trowel in a bin'' is unsuccessful because it does not recognize that \textit{the trowel} is indeed in \textit{a bin}, perhaps because it is relatively small in the image and does not contrast from the background. In Example H, LLaVA incorrectly decides that the procedure ``Put the bottle in the cabinet'' is successful because it hallucinates that a \textit{bottle} is in \textit{the cabinet}, despite neither object appearing in the image. The ability of this analysis to easily identify failures of visual perception in VLMs again may be useful for future work in this area.

The colors of these points can be used to characterize common behaviors of VLMs. Additional insights toward the fine-grained strengths and weaknesses of various approaches may be gained from analyzing these results by mistake type, or verb and noun categories.